%% file: main.tex
\documentclass{article}
\usepackage{iclr2027_conference,times}
\iclrfinalcopy

\input{sections/00-preamble}
\input{sections/00-title}

\author{Kushal Chakrabarti\\
  South Park Commons\\
  \texttt{kushalc@obviouslywrong.org}
}

\begin{document}
\maketitle

\input{sections/00-abstract}

\input{assets/fig_hero}
\input{sections/01-introduction}
\input{sections/02-theory}
\input{sections/03-field}
\input{sections/04-lab}
\input{sections/05-related}
\input{sections/06-discussion}

\input{sections/08-llm-usage}
\input{sections/09a-ethics}
\input{sections/09b-reproducibility}

\bibliography{references}
\bibliographystyle{iclr2027_conference}

\appendix
\raggedbottom  % appendix tables are [H] (inline): leftover space goes to the page foot, not between paragraphs
\input{sections/07-limitations}

\input{sections/10a-appendix-observational}

\input{sections/10b-appendix-inverse-ifeval}

\input{sections/10d-appendix-live}

\end{document}

%% file: sections/00-preamble.tex
\usepackage{hyperref}
\usepackage{url}
\usepackage{xcolor}
\usepackage[T1]{fontenc}
\usepackage[utf8]{inputenc}
\usepackage{microtype}
\usepackage[varqu]{zi4}
\usepackage{amsmath, amssymb}
\usepackage{graphicx}
\usepackage{subcaption}
\usepackage{booktabs}
\usepackage{enumitem}   % [nosep,leftmargin=*] contributions list in sections/01-introduction.tex
\usepackage{array}      % >{\raggedright...} column specs in the generated tables
\usepackage{tabularx}   % full-width X columns, same tables
\usepackage{adjustbox}  % shrink-only `max width` on wide generated tables (table_frailty, _verifiers)
\usepackage{placeins}   % \FloatBarrier in sections/09b-reproducibility.tex
\usepackage{float}      % [H]: appendix tables print where they are input (paper/scripts/table_*.py for 10a/10b)
\usepackage{tikz}
\usetikzlibrary{positioning}

\renewcommand{\sectionautorefname}{Section}
\renewcommand{\subsectionautorefname}{Section}
\renewcommand{\subsubsectionautorefname}{Section}
\DeclareRobustCommand{\apxref}[1]{{%
    \renewcommand{\sectionautorefname}{Appendix}%
    \renewcommand{\subsectionautorefname}{Appendix}%
    \renewcommand{\subsubsectionautorefname}{Appendix}%
    \autoref{#1}%
}}

\makeatletter
\newcommand{\pct}{\edef\cr@series{\f@series}\edef\cr@medium{m}%
\ifx\cr@series\cr@medium\else\kern-0.06em\fi\%}
\makeatother
\newcommand{\pp}{\ifmmode\mathrm{pp}\else pp\fi}

\input{assets/macros.tex}

\hypersetup{colorlinks=true, linkcolor=blue!50!black, citecolor=blue!50!black, urlcolor=blue!50!black}

\newenvironment{herofigure}{
\begin{figure}[h!]}{
\end{figure}}
\newcommand{\heropanelwidth}{0.49\textwidth}
\newenvironment{panelfigure}{
\begin{minipage}[t]{0.485\textwidth}}{
\end{minipage}}
\newcommand{\panelpairbegin}{
\begin{figure}[t]}
  \newcommand{\panelpairsep}{\hfill}
  \newcommand{\panelpairend}{
\end{figure}}
\newenvironment{captionedtable}[2]{\caption{#2}\label{#1}}{}
\makeatletter
\g@addto@macro\appendix{%
  \setcounter{topnumber}{3}%
  \setcounter{totalnumber}{4}%
}
\makeatother

%% file: assets/macros.tex
\newcommand{\RecallAgeSlopeCI}{-0.059 \pm 0.012}
\newcommand{\RecallRepos}{1,095}
\newcommand{\RecallBands}{3}
\newcommand{\RecallBandsNegative}{3}
\newcommand{\RecallOwnAuthor}{1.33}
\newcommand{\RecallOwnAuthorCI}{[1.09, 1.63]}
\newcommand{\RecallOwnSection}{1.16}
\newcommand{\RecallOwnSectionCI}{[1.03, 1.30]}
\newcommand{\RecallSectionStrata}{803}

\newcommand{\AccessT}{15}
\newcommand{\AccessWorlds}{64}
\newcommand{\AccessSeeds}{3}
\newcommand{\AccessK}{2}
\newcommand{\AccessL}{1}
\newcommand{\CoverDirectives}{300}
\newcommand{\CoverLooCost}{65.4\pp{}}
\newcommand{\CoverLooZ}{23.5}
\newcommand{\CoverExcessShare}{17.3\pct}
\newcommand{\CoverDstarSat}{85.6\pct}
\newcommand{\CoverVerifiedSat}{85.4\pct}
\newcommand{\CoverEmptySat}{22.2\pct}
\newcommand{\CoverVerifiedExcess}{$-$16.4\pct{}}
\newcommand{\AccessNoneExcessV}{+56.0\pct{}}

\newcommand{\AccessNoiseExcessV}{$-$0.2\pct{}}
\newcommand{\AccessNoiseSat}{37.8\pct}
\newcommand{\AccessCommitExcessV}{+12.7\pct{}}
\newcommand{\AccessCommitSat}{43.0\pct}
\newcommand{\AccessWriteForfeitPct}{53.6\pct}

\newcommand{\AccessCommitVsWriteZ}{2.29}

\newcommand{\AccessNoiseSatPp}{5.1\pp{}}
\newcommand{\AccessNoiseSatZ}{3.05}
\newcommand{\ObsFailureStated}{1.5\pct}
\newcommand{\ObsFailureStatedCI}{[1.0\pct, 2.1\pct]}
\newcommand{\ObsFailureCoded}{100}
\newcommand{\ObsRationaleRewriteHR}{0.67}
\newcommand{\ObsRationaleRewriteHRCI}{[0.57, 0.79]}
\newcommand{\ObsRationaleRewriteLessPct}{33\pct}
\newcommand{\ObsRewriteEvents}{1,077}
\newcommand{\ObsRewriteCommits}{1,018}
\newcommand{\ObsLengthSubjects}{5.0\pct}
\newcommand{\ObsReasonPhrase}{0.9\pct}
\newcommand{\ObsReasonPhraseCI}{[0.8\pct, 1.0\pct]}
\newcommand{\ObsReaddRewrite}{60\pct}
\newcommand{\ObsReaddRewriteCI}{[50\pct, 69\pct]}

\newcommand{\ObsSlopeNovel}{$-$5.9\pp{}}
\newcommand{\ObsSlopeNovelCI}{[-6.7, -5.1]}
\newcommand{\ObsSlopeRelearned}{+1.7\pp{}}
\newcommand{\ObsSlopeRelearnedCI}{[+1.0, +2.5]}
\newcommand{\ObsAdditions}{125,941}
\newcommand{\WorldRewritesControl}{0.41}
\newcommand{\WorldRewritesCommit}{0.14}
\newcommand{\WorldRewritesFold}{2.9}
\newcommand{\WorldRewritesDelta}{+0.27}
\newcommand{\WorldRewritesDeltaCI}{[+0.11, +0.45]}
\newcommand{\WorldRewritesDeltaP}{0.002}
\newcommand{\WorldRewritesGrewControl}{79\pct}

\newcommand{\WorldRewritesGrewDelta}{17\pp{}}
\newcommand{\WorldRewritesGrewDeltaCI}{[+7, +26]}
\newcommand{\CodeifNamedWorlds}{32}
\newcommand{\CodeifNamedSizePp}{$-$26.8\pp{}}
\newcommand{\CodeifNamedSatPp}{$-$2.9\pp{}}

\newcommand{\OutcomeOverprunedShare}{2.6\pct}

\newcommand{\BeyondWorlds}{32}
\newcommand{\BeyondT}{6}
\newcommand{\BeyondGapLow}{62.6\pp{}}
\newcommand{\BeyondGapHigh}{291.2\pp{}}
\newcommand{\BeyondSatLow}{$-$0.9\pp{}}
\newcommand{\BeyondSatHigh}{+7.4\pp{}}
\newcommand{\BeyondCodeNoneExcess}{+1.7\pct{}}

\newcommand{\BeyondCodeCommitExcess}{$-$32.7\pct{}}

\newcommand{\BeyondWildNoneExcess}{+38.5\pct{}}

\newcommand{\BeyondWildCommitExcess}{+14.5\pct{}}

\newcommand{\LiveFiles}{11}
\newcommand{\LiveTasks}{37}
\newcommand{\LiveMaxDirectives}{200}

\newcommand{\LiveDeltaSatPct}{17.7\pct}
\newcommand{\LivePostSatLoss}{4.5\pp{}}

\newcommand{\LiveCommitDropped}{17.1\pct}
\newcommand{\LiveControlDropped}{6.0\pct}

\newcommand{\LivePostShrink}{75.4\pct}

\newcommand{\LiveEvents}{387}
\newcommand{\LiveEventRepos}{280}

\newcommand{\LiveJointPairs}{680}
\newcommand{\LiveJointGap}{6.6\pp{}}
\newcommand{\LiveJointP}{0.003}
\newcommand{\LiveJointPostGap}{6.7\pp{}}
\newcommand{\LiveJointPostP}{0.001}

\newcommand{\LiveUnsignedWorlds}{6}
\newcommand{\LiveUnsignedPostGap}{7.2\pp{}}
\newcommand{\LiveUnsignedPostP}{0.031}
\newcommand{\LiveJointAuditItems}{25}
\newcommand{\LiveJointAuditLuna}{22}
\newcommand{\LiveJointAuditSecond}{2}
\newcommand{\FieldRepos}{1,867}
\newcommand{\FieldMultiVersionFiles}{1,801}
\newcommand{\FieldSpells}{247,694}
\newcommand{\FieldDeletions}{40,355}
\newcommand{\FieldRewriteDeaths}{84,233}
\newcommand{\FieldMigrationDeaths}{682}
\newcommand{\FieldRewriteToDeletionRatio}{2.1}

\newcommand{\FieldMultiVersionRepos}{1,576}

\newcommand{\FieldFracGrow}{64.3\pct}
\newcommand{\FieldFracShrink}{26.6\pct}
\newcommand{\FieldGrowthCountPeakPct}{+226\pct{}}

\newcommand{\FieldGrowthLenFiles}{1,776}
\newcommand{\FieldGrowthPayloadPct}{+140\pct{}}
\newcommand{\RewriteDeathShare}{67.2\pct}

\newcommand{\RewriteEventFiles}{52}
\newcommand{\RewriteEventWindow}{10}
\newcommand{\RewriteCutPct}{73.1\pct}
\newcommand{\RewriteNetCutPct}{60.7\pct}
\newcommand{\RewriteGrowthPreCommit}{2.8\pct}
\newcommand{\RewriteGrowthPostCommit}{4.8\pct}

\newcommand{\FieldNetPerCommit}{+4.7}
\newcommand{\FieldTouchingCommits}{19,267}
\newcommand{\FieldOpMedian}{39}
\newcommand{\FieldOpNinety}{131}
\newcommand{\FieldSlopeCommit}{-0.043}
\newcommand{\FieldSlopeCommitCI}{-0.043 \pm 0.011}

\newcommand{\HazardGeoExcess}{0.072}

\newcommand{\HazardGeoDecayCI}{0.222 \pm 0.024}

\newcommand{\HazardGeometricRsq}{0.815}
\newcommand{\HazardFrailtyRsq}{-0.482}

\newcommand{\HazardGeoSurvival}{74.5\pct}
\newcommand{\HazardGeoHalfLife}{3.1}
\newcommand{\CollapseRepos}{284}
\newcommand{\CollapseDeletions}{15,574}

\newcommand{\FieldFEFiles}{1,104}

\newcommand{\FrailtyAbsorbedFile}{4.8\pct}

\newcommand{\FlowGammaDeletionCI}{0.74 \pm 0.13}

\newcommand{\FlowGammaAdditionCI}{0.14 \pm 0.06}
\newcommand{\HorizonWorlds}{184}
\newcommand{\HorizonSat}{44.0\pct}
\newcommand{\HorizonControlExcessPct}{+211.3\pct{}}
\newcommand{\HorizonPracticalExcessPct}{+1.4\pct{}}
\newcommand{\HorizonT}{51}

\newcommand{\HorizonExcessRemovedPct}{99.3\pct}
\newcommand{\CapacityTiers}{3}
\newcommand{\CapacityControlRange}{+67.7\pct{} to +571.9\pct{}}
\newcommand{\AblationFreeformGivebackPct}{99.4\pct}

\newcommand{\LabVerifierTypes}{20}
\newcommand{\LabCommentCap}{1{,}024}

\newcommand{\SweepCells}{5}

\newcommand{\SweepRewriteCells}{3}

\newcommand{\SweepRegrowthFasterCells}{2}
\newcommand{\SweepFormHeldCells}{5 of 5}
\newcommand{\SweepFormMarginMin}{0.280}
\newcommand{\SweepGeoDecayRange}{0.200 to 0.222}

\newcommand{\SegMinWords}{2}
\newcommand{\MatchFuzzyThreshold}{70}
\newcommand{\RewriteFracThreshold}{75.0\pct}
\newcommand{\RewriteMinDirectives}{5}
\newcommand{\HazardMinDuration}{3}
\newcommand{\HazardBandwidthCommits}{2}

\newcommand{\HazardBoot}{200}
\newcommand{\HazardAgeMax}{50}

\newcommand{\GateBPrecision}{1.000}
\newcommand{\GateBRecall}{0.933}
\newcommand{\GateBFloor}{0.85}
\newcommand{\FieldMatches}{299,440}
\newcommand{\InlineExposure}{85.1\pct}

\newcommand{\InlineWorlds}{184}

\newcommand{\InlineDeltaExcessCI}{$28.6 \pm 26.5$\pp{}}

\newcommand{\InlineDeltaSatCI}{$-1.7 \pm 3.4$\pp{}}
\newcommand{\InlineExcessPct}{+28.5\pct{}}
\newcommand{\InlineRefExcessPct}{$-$0.1\pct{}}
\newcommand{\CostCalls}{49,680}
\newcommand{\CostTokens}{37.9M}
\newcommand{\CostModelHours}{45}
\newcommand{\CostMaintainerInRange}{7.0M to 9.9M}

%% file: sections/00-title.tex
\title{Why Does \texttt{CLAUDE.md} Keep Growing?\\ Catastrophic Remembering in Agentic Coding}

%% file: sections/00-abstract.tex
\begin{abstract}
  Agentic prompts like \texttt{CLAUDE.md} grow without bound in real repositories, stopping only when
  history ends or someone rewrites the file wholesale.
  We trace this to imperfect recall: appending an instruction is always cheap, but once an instruction's
  latent reasoning is gone, deleting it without risking a regression costs $O(2^{|D|})$ in a prompt of
  $|D|$ instructions.
  Successive edits destroy that latent reasoning, additions outrun deletions and prompts
  grow until agents become non-functional --- \textbf{catastrophic remembering}.
  First, we characterize this phenomenon across \FieldSpells{} instruction lifetimes in \FieldRepos{} repositories:
  agentic prompts grow nearly monotonically, more than tripling over their lifetime (\FieldGrowthCountPeakPct{})
  at \FieldNetPerCommit{} net instructions per commit. Deletion hazard decays geometrically with
  instruction age, consistent with imperfect recall and with no other rival we test.
  Then, we show that prompt comments can halt the growth: in verifiable worlds built by inverting IFEval,
  \textbf{comments encoding latent reasoning remove \HorizonExcessRemovedPct{} of excess size}
  (\HorizonControlExcessPct{} to \HorizonPracticalExcessPct{}).
  Finally, we show comments yield smaller real-world and coding-convention prompts than existing
  context maintainers, grow more effective with model capability and, at small scale,
  \textbf{improve instruction-following in real repositories by
  \LiveDeltaSatPct{}}.
  If English is the new code, why don't we have comments yet?
\end{abstract}

%% file: assets/fig_hero.tex
\begin{herofigure}
  \centering
  \begin{subfigure}[t]{\heropanelwidth}
    \centering
    \includegraphics[width=\linewidth]{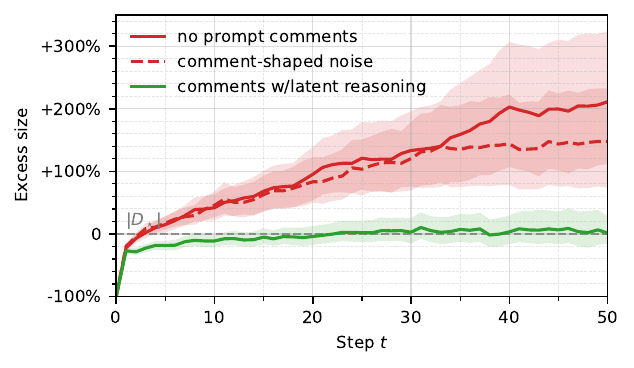}
    \caption{
      \textbf{Causal:} Prompts without comments (or with comment-shaped noise) grow without bound; when
      comments carry latent reasoning, they end smallest at equal constraint satisfaction. Excess size is
      $\frac{\lvert D_t \rvert}{\lvert D_\star \rvert}{-}1$ for IFEval reference prompt $D_\star$;
      bands are 95\% bootstrap over \HorizonWorlds{} worlds.
    }
    \label{fig:hero:lab}
  \end{subfigure}
  \panelpairsep
  \begin{subfigure}[t]{\heropanelwidth}
    \centering
    \includegraphics[width=\linewidth]{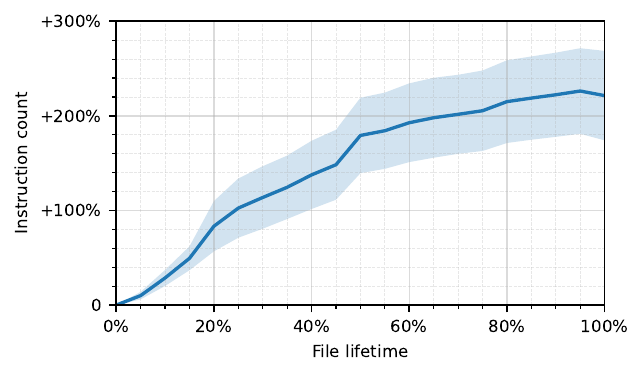}
    \caption{
      \textbf{Observational:} After creation, the number of instructions in an agentic prompt
      more than triples (\FieldGrowthCountPeakPct{}) over the file's lifespan, across \FieldRepos{} GitHub
      repositories and \FieldSpells{} instruction lifetimes. Instruction count relative to file's first
      version, $\frac{\lvert D_t \rvert}{\lvert D_0 \rvert}{-}1$; lifetime normalized per file.
    }
    \label{fig:hero:growth}
  \end{subfigure}
  \caption{
    Deleting prompt instructions requires remembering why they were added; agentic prompts lack comments, so
    they only grow over their lifetime.
  }
  \label{fig:hero}
\end{herofigure}

%% file: sections/01-introduction.tex
\input{assets/fig_diagram.tex}

\section{Introduction}
\label{sec:intro}

Agentic prompts almost never stop growing. A \texttt{CLAUDE.md}, \texttt{AGENTS.md}, or
\texttt{copilot-instructions.md} gains instructions and rarely sheds them \citep{chatlatanagulchai2025agentreadmes}.
Deleting the file is not the answer: agents in repositories carrying one run faster and emit fewer output tokens
\citep{lulla2026contextfiles}. Keeping everything is not free either: instruction-following degrades as
constraints accumulate \citep{jiang2024followbench, lior2025wildifeval}, and the median file already carries
\FieldOpMedian{} instructions, the average more than tripling over its own lifetime
(\FieldGrowthCountPeakPct{}, \autoref{fig:hero:growth}).

Different mechanisms can explain this growth.
If requirements stop applying (instruction staleness), deletion hazard rises with age. If brittle
instructions die quickly (content fragility) or fixed effort gets spread across ever-growing files,
compositional effects dominate.
On the other hand, if maintainers lose their latent reasoning (imperfect recall), hazard falls with age.
Line diffs and file sizes cannot resolve individual instructions over time, so the mechanistic question has
remained open.

Tracking individual instructions answers it. \RewriteDeathShare{} of deaths arrive in one
file-bulldozing commit, after which growth resumes: removing one instruction needs a reason, removing all
needs none. The hazard agrees, decaying geometrically with age (\autoref{sec:field}).

``What does this code do again?'' is a question every engineer has asked at least once.
Deleting an agentic instruction risks a regression, and doing so safely requires counterfactuals maintainers
cannot run.
Recording an instruction's latent reasoning costs $O(1)$, but reconstructing it costs $O(2^{|D|})$ for a
prompt with $|D|$ instructions (\autoref{sec:theory}).
Successive edits destroy that latent reasoning, so deletion hazard decays with age, additions outrun
removals, and instructions grow until agents become non-functional --- \textbf{catastrophic remembering}.
Under catastrophic forgetting, a gradient learner overwrites what it should have kept
\citep{mccloskey1989catastrophic, french1999catastrophic, kirkpatrick2017overcoming}; under catastrophic
remembering, a maintainer keeps what they should have overwritten.
Both fail and fail for the same reason: whatever would license the update is gone.\looseness=-1

Software engineering solved this decades ago with the comment: text addressed to the next maintainer and
invisible to the interpreter \citep{knuth1984literate}. Recovering why code was written as it was is
developers' most serious problem \citep{latoza2006mentalmodels}, so the practice is to record the why, not
the how or the what \citep{mcconnell2004codecomplete}. Unsurprisingly, problem and practice carry over to
agentic prompts.

We show that prompt comments encoding an instruction's latent reasoning can halt unbounded prompt growth,
removing \HorizonExcessRemovedPct{} of the excess size (\HorizonControlExcessPct{} to
\HorizonPracticalExcessPct{}) over \HorizonT{} steps (\autoref{fig:hero:lab}).
Comments also keep real-world and coding-convention prompts smaller than existing context
maintainers do, and because irrelevant context distracts models \citep{shi2023distracted}, comment-enabled
pruning of real repositories' agentic prompts improves constraint satisfaction by \LiveDeltaSatPct{} at small scale
(\autoref{sec:wild}, \autoref{sec:live}).
Ablating what a comment records shows the gain needs latent reasoning tied to observed outcomes. Finally, the advantage
grows with maintainer capability (\autoref{sec:lab}).\looseness=-1

We contribute a theory, a field signature, an evaluation and a fix for prompt maintenance:
\begin{itemize}[nosep,itemsep=2pt,parsep=2pt,topsep=2pt,leftmargin=*]
  \item \emph{Theory of prompt maintenance.} Decaying recoverability of latent reasoning predicts unbounded
    growth (\autoref{sec:theory}).
  \item \emph{Signature of prompt growth.} Deletion hazard decays geometrically with age over
    \FieldSpells{} lifetimes, controlling for file size, consistent with imperfect recall and no
    other tested rival
    (\autoref{sec:hazard}).
  \item \emph{New evaluation with verified optimum.} Inverting IFEval yields verifiable worlds whose
    minimum covers are verifiable, so excess size and constraint satisfaction become measurable
    (\autoref{sec:testbed}).
  \item \emph{Practical solution.} Prompt comments remove \HorizonExcessRemovedPct{} of the excess size
    (\autoref{sec:protocol}), and at small scale improve constraint satisfaction in real
    repositories by \LiveDeltaSatPct{} (\autoref{sec:live}).
\end{itemize}

%% file: assets/fig_diagram.tex
\begin{figure*}[t]
  \centering
  \begin{subfigure}[b]{0.58\textwidth}
    \centering
    \resizebox{\linewidth}{!}{\input{assets/fig_flow}}
    \caption{
      Why do agentic prompts only grow? The latent reasoning behind $d_i$ is never written and decays over
      the (possibly agentic) development loop, so deletion risks an acute regression while the cost of
      addition stays diffuse.
    }
    \label{fig:flow}
  \end{subfigure}\hfill
  \begin{subfigure}[b]{0.38\textwidth}
    \centering
    \fbox{\parbox{0.96\linewidth}{\input{assets/fig_comment}}}
    \caption{
      The fix: comments carrying latent reasoning encode why an instruction was added, which stops
      unbounded growth in agentic prompts.
    }
    \label{fig:comment}
  \end{subfigure}
  \caption{
    The prompt-comment world. Latent reasoning is free at write-time and exponentially costly to reconstruct
    at read-time, but prompt comments preserve it and stop the unbounded growth of agentic prompts.
  }
  \label{fig:comment_world}
\end{figure*}

%% file: assets/fig_flow.tex
\begin{tikzpicture}[
    >=stealth,
    font=\sffamily,
    % Core styles: uniform standard font size, 1.6pt line widths (100% thicker)
    stage_title/.style={font=\sffamily\bfseries, text=black!80},
    mem/.style={draw=black!60, line width=1.6pt, rounded corners=1mm, fill=blue!5, align=center, inner sep=2.5mm},
    % why box explicitly defines the 2.5mm intra-box margin
    why/.style={draw=red!60, line width=1.6pt, rounded corners=1.5mm, fill=red!5, align=center, inner sep=2.5mm},
    ann/.style={font=\sffamily\bfseries, text=red!80!black, align=center},
    flow/.style={->, line width=1.6pt, draw=black!70}
  ]

  % Bounding box mathematically locked to 2.81 aspect ratio to prevent distortion (14 / 4.98)
  % Y=0.0 is precisely calibrated to match the header/box margin
  \useasboundingbox (0, 0) rectangle (14, 4.98);

  % ---------------------------------------------------------
  % PLANE 1: TEXT / TITLES (Y = 4.4)
  % ---------------------------------------------------------
  \node[stage_title] at (2.25, 4.4) {WRITE ($t=0$)};
  \node[stage_title] at (7.625, 4.4) {THE LOOP ($t=1..k$)};
  \node[stage_title] at (12.5, 4.4) {EDIT ($t=k$)};

  % ---------------------------------------------------------
  % PLANE 2: OBSERVABLES (Y = 3.2)
  % ---------------------------------------------------------
  \node[mem] (fail) at (1.0, 3.2) {fail: \textcolor{red}{\bfseries $\times$}};
  \node[mem] (d4) at (3.5, 3.2) {$d_4$};
  \node[mem, text width=1.8cm] (loopm) at (7.625, 3.2) {$d_1 \dots d_3$ \\ \vspace{2mm} $+d_4 \dots d_6$};
  \node[mem, text width=1.8cm] (readm) at (12.5, 3.2) {$d_1 \dots d_3$ \\ \vspace{2mm} $d_4 \dots d_6$};

  % Observable execution flows
  \draw[flow] (d4) -- (loopm);
  \draw[flow] (loopm) -- (readm);

  % ---------------------------------------------------------
  % PLANE 3: HIDDEN REASONING (Y = 1.7)
  % ---------------------------------------------------------
  \node[why] (why) at (2.25, 1.7) {why?};

  % Caption margin mathematically locked to exactly match the 2.5mm intra-box margin above
  \node[ann, anchor=north] at ([yshift=-2.5mm]why.south) {latent reasoning \\ never written};

  % Right-angled causal pathways
  \draw[flow] (fail.south) |- (why.west);
  \draw[flow] (why.east) -| (d4.south);

  % Depictive execution loop reaching deep into the reasoning plane (down to Y=1.2)
  \draw[flow] ([xshift=-0.7cm]loopm.south)
  to[out=270, in=180] (7.625, 1.2)
  to[out=0, in=270] ([xshift=0.7cm]loopm.south);

  % ---------------------------------------------------------
  % SMOOTH MANUAL DEGRADATION (X = 3.5 to X = 9.75)
  % ---------------------------------------------------------
  % Originates at the right-angle corner of the why->d4 arrow
  \draw[line width=1.6pt, draw=black!80] (3.5, 1.7) -- (5.25, 1.7); % Solid up to Write/Loop boundary

  % Exponentially decaying segment length, increasing gaps, and fading opacity
  \draw[line width=1.6pt, draw=black!75] (5.25, 1.7) -- (5.55, 1.7);
  \draw[line width=1.6pt, draw=black!70] (5.65, 1.7) -- (5.90, 1.7);
  \draw[line width=1.6pt, draw=black!65] (6.02, 1.7) -- (6.22, 1.7);
  \draw[line width=1.6pt, draw=black!60] (6.36, 1.7) -- (6.52, 1.7);
  \draw[line width=1.6pt, draw=black!55] (6.68, 1.7) -- (6.81, 1.7);
  \draw[line width=1.6pt, draw=black!50] (6.99, 1.7) -- (7.09, 1.7);
  \draw[line width=1.6pt, draw=black!45] (7.29, 1.7) -- (7.37, 1.7);
  \draw[line width=1.6pt, draw=black!40] (7.60, 1.7) -- (7.66, 1.7);
  \draw[line width=1.6pt, draw=black!35] (7.92, 1.7) -- (7.96, 1.7);
  \draw[line width=1.6pt, draw=black!30] (8.25, 1.7) -- (8.28, 1.7);
  \draw[line width=1.6pt, draw=black!25] (8.60, 1.7) -- (8.62, 1.7);
  \draw[line width=1.6pt, draw=black!20] (8.97, 1.7) -- (8.98, 1.7);
  \draw[line width=1.6pt, draw=black!15] (9.35, 1.7) -- (9.36, 1.7);
  \draw[->, line width=1.6pt, draw=black!10] (9.70, 1.7) -- (9.75, 1.7);

  % ---------------------------------------------------------
  % VERTICAL DIVIDERS AND BOUNDARY MARKERS
  % ---------------------------------------------------------
  \draw[dashed, black!20, line width=1.6pt] (5.25, 4.9) -- (5.25, 0.0);
  \draw[dashed, black!20, line width=1.6pt] (10.0, 4.9) -- (10.0, 0.0);

  % Degradation failure point exactly at the boundary (X=10.0)
  \node[text=red, font=\Huge\bfseries, inner sep=0pt] at (10.0, 1.7) {$\times$};

  % Read-time reasoning void
  \node[text=red, font=\Huge\bfseries] at (12.5, 1.7) {?};

\end{tikzpicture}

%% file: assets/fig_comment.tex
\definecolor{synKey}{RGB}{135,0,175}
\definecolor{synMark}{RGB}{0,92,197}
\definecolor{synComment}{RGB}{0,128,0}
{\small\linespread{0.85}\selectfont\ttfamily\raggedright
  \textcolor{gray}{
    Write a brief post about how Earth is different from Venus.\\
    ...\\
    Write your response as three sentences, each ending with a period.
  }\textcolor{synComment}{\# Task 0 continues to fail on sentence count; last instruction was
  vague; maybe being explicit about periods will help?}\par
}

%% file: sections/02-theory.tex
\section{Forgetting Requires Remembering Why}
\label{sec:theory}

\input{assets/fig_mass_rewrite.tex}

We model prompt maintenance as the \emph{online estimation of an unobservable constraint set from
censored, noisy feedback}. We posit that (i) deleting an instruction requires its \emph{latent reasoning}, the
original reason that a maintainer wrote it, and (ii) that latent reasoning decays with age. If it decays
geometrically, a fixed fraction of every instruction ever written is never deleted, and the prompt grows
linearly without bound. \autoref{sec:field} tests the geometric form.

A task $j = (o_j, C_j)$ pairs a stated objective $o_j$ with its unobservable constraints $C_j \subseteq C$,
drawn from a fixed hidden set $C = \{g_c\}$ of verifiers. The prompt at time $t$ is a finite set of
instructions $D_t$; instruction $d$ enters at time $\tau_d$ and has age $a_d = t - \tau_d$. A response to
$j$ satisfies $c \in C_j$ with probability $q_c(D, j)$, and $C$ is observed only through those draws. At
each timestep a (possibly agentic) \emph{maintainer} updates $D_t$, and the \emph{harness} draws tasks, which
the agentic \emph{executor} solves under $D_t$; finally, the harness verifies the results against $C_j$.

The maintainer's target is the \textbf{minimum cover} $D_\star$, the smallest instruction set that maximizes
expected constraint satisfaction:
\begin{equation}
  \begin{split}
    D_\star &= \operatorname*{argmin}_{D \in \mathcal{M}} |D|, \\
    \mathcal{M} &= \operatorname*{argmax}_{D}\, \mathrm{E}_{j,\,c \in C_j}[q_c(D, j)]
  \end{split}
\end{equation}
Every experiment in this paper scores $D_t$ against $D_\star$ on two axes: \emph{constraint satisfaction}, the
expected $\mathrm{E}[q_c]$, and \emph{excess size}, $\frac{|D_t|}{|D_\star|} - 1$. Maintenance regimes
win by improving one axis without degrading the other.\looseness=-1

We assume only that $q_c$ meets the following properties:
\input{assets/table_characteristics.tex}
Properties A1–A3 (instructability, interference, and redundancy) set the
overall regime whereas properties A4 (censoring) and A5 (stochasticity) respectively block deletions and
drive additions.\looseness=-1

Deleting without risking a regression requires an infeasible number of counterfactuals.
Write $d$'s contribution to $c$ as $\Delta_c(d \mid D) = \mathbb{E}_j[q_c(D, j) - q_c(D \setminus \{d\}, j)]$
over tasks with $c \in C_j$; $d$ is \emph{excess} when $\Delta_c \le 0$ for every $c$.
Estimating it means probing $D \setminus \{d\}$ but redundancy can defeat individual probes: multiple
instructions covering a single constraint can individually look free under ablation but, if they're deleted
jointly, causes material degradation.
An honest audit thus costs $O(2^{|D_t|})$ subset probes, and $D_\star$ is incomputable in general. Knowing
why the maintainer added $d$ --- its latent reasoning $r_d$ --- collapses that to $O(1)$, but $r_d$ decays
with $a_d$.\looseness=-1

\panelpairbegin
\input{assets/fig_rewrite_ratchet.tex}\panelpairsep
\input{assets/fig_hazard.tex}
\panelpairend

Deletion hazard is defined by the recoverability of an instruction's latent reasoning. Define the decay
\begin{equation}
  \rho(d, a) = \Pr(r_d \text{ recoverable at age } a),
\end{equation}
with marginal $\bar \rho(a) = \mathbb{E}_d\,\rho(d, a)$, so that $\bar \rho(0) = 1$ and $\bar \rho(a) \to 0$
as $a \to \infty$. We call that decay \textbf{imperfect recall}: the instruction stays and its failures
recur, but its latent reasoning becomes increasingly unrecoverable. A maintainer rationally deletes only when
the latent reasoning survives and the instruction verifies as excess, so the hazard in its own age factorizes,
\begin{equation}
  h(a) \;\approx\; \bar \rho(a)\, s(a),
  \label{eq:hazard}
\end{equation}
for $s(a) = \Pr(\text{excess} \mid a)$, the event-history form used for written organizational rules
\citep{schulz1998limits, march2000dynamics, zhou1993dynamics}. We take the forgetting curve to be geometric,
$\bar\rho(a) =
e^{-ba}$, for forgetting rate $b$ and a latent-reasoning half-life of $\ln 2 / b$ commits, and leave the excess
process free: everything below needs only $s(a) \le 1$.

An instruction survives to age $a$ with probability $S(a) = \prod_{k \le a} (1 - h(k))$. With $A$
additions per commit and deletions independent across instructions,
\begin{equation}
  \mathbb{E}\,|D_t| \;=\; A \sum_{a < t} S(a).
  \label{eq:size}
\end{equation}
The prompt settles iff the expected lifetime $\mathbb{E}[L] = \sum_a S(a)$ is finite, at $|D_\infty| =
A\,\mathbb{E}[L]$, which requires $\sum_a h(a) = \infty$. Under geometric decay the sum converges, whatever the
excess process: $\sum_a h(a) \le \sum_a e^{-ba} = 1/(1 - e^{-b}) < \infty$, so $S_\infty = \prod_k (1 - h(k)) > 0$ and
\begin{equation}
  \mathbb{E}\,|D_t| \;=\; A\, S_\infty\, t + O(1).
  \label{eq:ratchet}
\end{equation}

\autoref{eq:ratchet} is catastrophic remembering in closed form: as latent reasoning is lost, a fraction
$S_\infty$ of every instruction ever written is never deleted, deletions per commit saturate at
$A(1 - S_\infty)$ while additions continue at $A$, and excess size grows linearly with no change in task or
constraints. Recoverability is thus necessary and, as \autoref{sec:protocol} shows, also sufficient for a
bounded prompt.

%% file: assets/fig_mass_rewrite.tex
\begin{figure*}[t]
  \includegraphics[width=\linewidth]{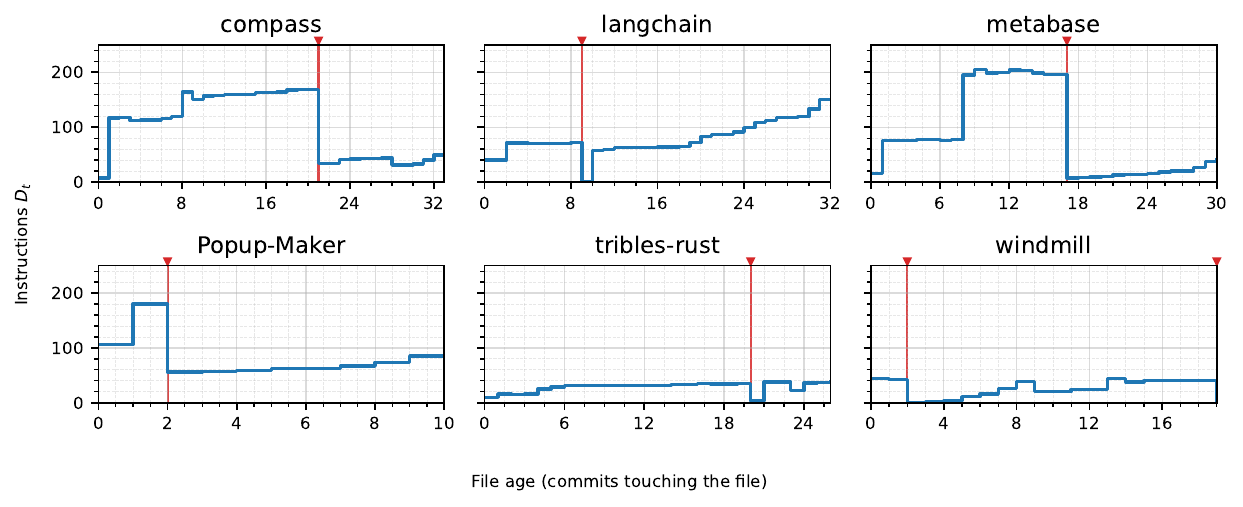}
  \caption{
    Growth, wholesale rewrite, regrowth --- agentic prompts almost never
    shrink incrementally. Instruction count $D_t$ for six illustrative
    agentic prompts; red rules mark commits cutting at least
    \RewriteFracThreshold{} of a file holding at least \RewriteMinDirectives{} instructions. Such rewrites carry
    \RewriteDeathShare{} of instruction deaths; the median file is cut by \RewriteCutPct{} and
    sits \RewriteNetCutPct{} below it after \RewriteEventWindow{} commits
    (\RewriteEventFiles{} files, \autoref{fig:massrewrite:event}).
  }
  \label{fig:massrewrite:examples}
\end{figure*}

%% file: assets/table_characteristics.tex
\begin{center}
  \small
  % Narrow label column: @{\hspace{0.6em}} trims the gap after it.
  \begin{tabular}{@{}l@{\hspace{0.6em}}ll@{}}
    \toprule
    & \textbf{Property} & \textbf{Behavior} \\
    \midrule
    A1 & instructability & some $d$ raises $q_c$ \\
    A2 & interference & adding $d$ lowers other $q_{c'}$ \\
    A3 & redundancy & a second $d'$ adds no $q_c$ \\
    A4 & censoring & outcomes don't name their $d$ \\
    A5 & stochasticity & $q_c < 1$ even when covered \\
    \bottomrule
  \end{tabular}
\end{center}

%% file: assets/fig_rewrite_ratchet.tex
\begin{panelfigure}
  \centering
  \includegraphics[width=\linewidth]{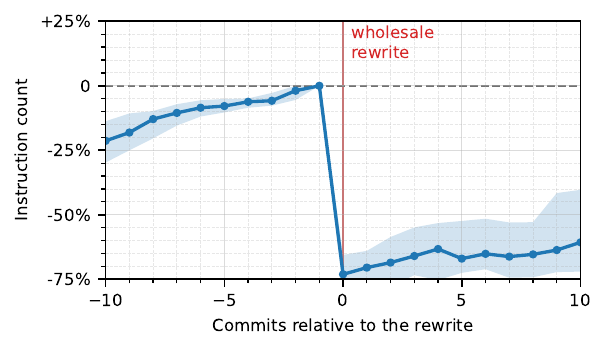}
  \caption{
    The ratchet: a rewrite resets a prompt's size but not its growth rate. Prompts grow
    \RewriteGrowthPreCommit{} per commit before their first rewrite, are cut by \RewriteCutPct{}, then
    regrow immediately and faster, at \RewriteGrowthPostCommit{} per commit. Median instructions
    relative to the \emph{first} pre-rewrite, each file's rewrite event aligned to $t=0$; band: 95\%
    bootstrap CI.
  }
  \label{fig:massrewrite:event}
\end{panelfigure}

%% file: assets/fig_hazard.tex
\begin{panelfigure}
  \centering
  \includegraphics[width=\linewidth]{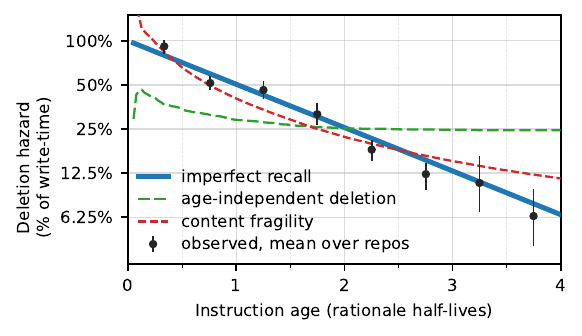}
  \caption{
    Deletion hazard halves repeatedly --- the imperfect-recall signature, which
    age-independent deletion (flat) misses everywhere and content fragility (bending) past two.
    Age and hazard are rescaled by each repository's own decay (\autoref{tab:hazardforms}).
    \CollapseRepos{} repositories observed to four half-lives, \CollapseDeletions{} deletions, 95\pct{} bootstrap.
  }
  \label{fig:hazard}
\end{panelfigure}

%% file: sections/03-field.tex
\section{Prompts Ratchet: Growth, Rewrite, Regrowth}
\label{sec:field}

Agentic prompts grow until someone rewrites them en masse, then grow again --- a sawtooth we call the
\textbf{ratchet} (\autoref{fig:massrewrite:event}). Its deletion hazard and covariates,
alongside the causal test in \autoref{sec:lab}, identify its root cause: imperfect recall.

We decompose agentic prompts into individual instructions and track their decay over commit histories
(\autoref{sec:growth}-\ref{sec:rewrite}), then test four rival mechanisms against the geometric hazard
derived in \autoref{sec:theory} (\autoref{sec:hazard}). Our corpus
spans \FieldRepos{} GitHub repositories and \FieldSpells{} instruction lifetimes; \apxref{sec:appendix:corpus}
details construction, segmentation, matching and censoring.

\subsection{Agentic prompts grow unbounded until a wholesale rewrite}
\label{sec:growth}

Maintainers almost never remove instructions, so agentic prompts grow without bound until history ends or a file
is rewritten wholesale.
Over its lifetime, a file's instruction count more than triples (\FieldGrowthCountPeakPct{},
\autoref{fig:hero:growth}), a net \FieldNetPerCommit{} instructions per commit outside mass rewrites.
Instruction length and repository-wide prompt text grow too, so the count is not text moving between files (\apxref{sec:appendix:decomposition}).

At its last tracked version, the median file holds \FieldOpMedian{} instructions,
well past the threshold at which instruction-following degrades \citep{jiang2024followbench,
lior2025wildifeval}.
Every excess instruction is another that executors must needlessly follow, inducing instruction-following
costs that \autoref{sec:wild} and \autoref{sec:live} measure directly.

\subsection{Most disappearances are not deletions}
\label{sec:measurement}

Most instructions vanish en masse, not one at a time.
\RewriteDeathShare{} of instruction deaths (\FieldRewriteDeaths{}) arrive in a \emph{rewrite}, a commit killing
at least \RewriteFracThreshold{} of a file's instructions.
Only \FieldMigrationDeaths{} more arrive in a \emph{migration} of text to a sibling file, where a line diff
cannot separate deletion from rewording.
Maintainers do not judge individual instructions in either event \citep{march2000dynamics}, so we censor both
as competing risks, leaving \FieldDeletions{} tracked deletions (\apxref{sec:appendix:censoring}).\looseness=-1

\subsection{Regrowth resumes faster after a rewrite}
\label{sec:rewrite}

A mass rewrite resets a prompt's size but not its growth: the prompt starts growing again afterwards.
Aligning \RewriteEventFiles{} files on their first rewrite at $t = 0$, the
median instruction count rises through $t = -1$, is cut by \RewriteCutPct{} at $t = 0$, and then regrows
\emph{faster} than before: +\RewriteGrowthPostCommit{} instructions per commit after against
+\RewriteGrowthPreCommit{} before (\autoref{fig:massrewrite:event}).

Recorded latent reasoning predicts fewer rewrites: only \ObsReasonPhrase{} of instructions
record it, yet files that do are rewritten \ObsRationaleRewriteLessPct{} less often (\apxref{sec:appendix:rationale}).

\subsection{Deletion hazard falsifies rival mechanisms}
\label{sec:hazard}

The deletion hazard $h(a)$ and its covariates are consistent with imperfect recall and with no other rival
we test.

The hazard decays geometrically. It falls with age within files as well as across them (slope
$\FieldSlopeCommitCI$ per commit; \apxref{sec:appendix:hazard}, \autoref{fig:hazard}). Fitting \autoref{eq:hazard}
within each repository and pooling
by random effects gives $b = \HazardGeoDecayCI$, a latent-reasoning half-life of
\HazardGeoHalfLife{} commits and, by \autoref{eq:ratchet}, \HazardGeoSurvival{} of instructions never deleted.
The geometric form fits at $R^2_w = \HazardGeometricRsq$, the best of the five hazard shapes we fit
(\autoref{tab:hazardforms}).

Two rivals predict the opposite sign. If requirements go stale (instruction staleness;
\citealp{lehman1980laws}), hazard should rise with age; it falls (\autoref{fig:hazard}). If files drift
to new concerns (relevance drift), instructions turn excess and hazard should rise; it falls, so drift
can only mask forgetting, not explain authors deleting their own instructions $\RecallOwnAuthor\times$
as often as others' (\apxref{sec:appendix:dilution}).

The other two get the direction right but not the size.
If fragile instructions die young (content fragility), a gamma frailty model \citep{vaupel1979frailty} should
fit; however, it fits worse than a constant hazard ($R^2_w = \HazardFrailtyRsq$), and repository-level frailty explains
only \FrailtyAbsorbedFile{}.
On the other hand, if maintainers' fixed efforts are spread over an ever-growing file (effort dilution),
deletions should stay flat as they grow; instead, they grow sublinearly (exponent $\FlowGammaDeletionCI$),
suggesting partial dilution. After controlling for file size, we see that hazard still falls with age
($\RecallAgeSlopeCI$ per commit; \apxref{sec:appendix:hazard}), so dilution cannot explain it all.

Although indicative, this evidence is correlational; we next causally test the mechanism
by assigning latent reasoning directly.

%% file: sections/04-lab.tex
\input{assets/table_access.tex}

\section{Comments Halt the Ratchet and Buy Back Instruction-Following}
\label{sec:lab}

Prompt comments can halt the unbounded growth of instructions in agentic prompts and, in doing
so, improve constraint satisfaction through smaller, more robust prompts.

We first build an evaluation whose minimum cover is known (\autoref{sec:testbed}), use it to show that
comment-enabled prompts end smallest at matched constraint satisfaction (\autoref{sec:protocol}), and then
carry the result to real-world prompts and coding conventions (\autoref{sec:wild}) and finally to real
repositories (\autoref{sec:live}).

\subsection{Inverting IFEval makes the minimum cover observable}
\label{sec:testbed}

A maintainer must learn a prompt that covers every unobservable constraint while staying minimal, since
excess instructions cost constraint satisfaction (\autoref{sec:intro}).
Scoring this process requires the minimum cover, which is incomputable in general (\autoref{sec:theory}). Inverting
IFEval \citep{zhou2023ifeval} makes it observable. For each benchmark item $(D_j, C_j)$ --- instructions $D_j$,
verifiers $C_j$ --- we apply the following transform:

\begin{enumerate}[nosep,itemsep=2pt,parsep=2pt,topsep=2pt,leftmargin=*]
  \item \emph{Hide $D_j$.} Its stated instructions become the reference cover $D_\star$, so excess
    size is measurable.
  \item \emph{Keep $C_j$.} Its verifiers become the world's hidden constraint set, run by the
    harness and never directly disclosed to the maintainer.
  \item \emph{Summarize $D_j$ as $o_j$.} A stronger model lossily summarizes $D_j$ as a
    general task objective (``draft a product announcement'').
\end{enumerate}

A fresh maintainer then iteratively reconstructs $D_j$ from $o_j$ over $T$ rounds, seeing only censored,
noisy feedback from $C_j$.
Each arm sets whether the maintainer may annotate an instruction $d$ with a comment $r_d$ carrying that
instruction's latent reasoning.
The next maintainer reads $D_t$ and every $r_d$; the executor reads only $D_t$.

  We also verify the IFEval-provided reference instructions.
  Evaluated directly under the executor, $D_\star$ satisfies \CoverDstarSat{} of hidden constraints against
  \CoverEmptySat{} under an empty prompt.
  A leave-one-out evaluation further identifies the \emph{minimum verified cover} $D_V$, which satisfies
  \CoverVerifiedSat{} at \CoverVerifiedExcess{} excess size, smaller than $D_\star$ (\autoref{tab:access}).
  Implementation details are in \apxref{sec:appendix:testbed}.

\input{assets/fig_comment_examples.tex}
\input{assets/table_capacity.tex}

\subsection{Comments yield minimal prompts at matched constraint satisfaction}
\label{sec:protocol}

Prompt comments encoding an instruction's latent reasoning yield the smallest prompts at matched constraint
satisfaction: over \HorizonT{} rounds, comment-enabled prompts end at \HorizonPracticalExcessPct{} excess size against
\HorizonControlExcessPct{} uncommented, both satisfying \HorizonSat{} of hidden constraints (\autoref{fig:hero:lab},
\autoref{tab:arms}).

  The mass rewrites of public GitHub repositories (\autoref{sec:rewrite}) also emerge unprompted in this
  setting (\apxref{sec:appendix:worldrewrites}), and prompt comments cut them \WorldRewritesFold{}$\times$:
  uncommented maintainers mass-rewrite \WorldRewritesControl{} times per world, against \WorldRewritesCommit{}
  for comment-enabled maintainers (paired $p = \WorldRewritesDeltaP$).
  As in the observational setting, prompts grow before rewrites: \WorldRewritesGrewControl{} had grown over the
  preceding three rounds.

Controls and ablations trace the gain to agentic maintainers writing and reading back its latent reasoning. Each comment
records the failure behind an instruction, the hypothesized root cause and solution, and its ongoing performance
(\autoref{fig:comment_examples}).
With one shared maintenance prompt for every arm, comment-enabled maintainers shrink the working
  prompt from \AccessNoneExcessV{} to \AccessCommitExcessV{} above $D_V$ at matched constraint satisfaction
  (\autoref{tab:access}).
  Maintainers that write comments but never read them forfeit \AccessWriteForfeitPct{} of that gain ($z =
  \AccessCommitVsWriteZ$); those that read unrelated comments shrink just
as far (\AccessNoiseExcessV{}) but lose \AccessNoiseSatPp{} of constraint satisfaction ($z = \AccessNoiseSatZ$).
Replacing the structured schema with a free-form narrative gives back \AblationFreeformGivebackPct{} improvement,
while removing subsequent performance or ruled-out hypotheses causes material degradation
(\apxref{sec:appendix:ablations}).

  Comments reach the minimal prompt within the bounds of what agentic maintainers can learn in our environment. No arm
  approaches the \CoverDstarSat{} satisfied by $D_\star$ (\autoref{tab:access}), a ceiling that we surmise is set by
  learning from censored feedback, not by comments.

  Disaggregating the maintainer's performance, we show that comments win at every final size. Grouping worlds
  by the comment-enabled prompt's final size (at, below or above
  $|D_\star|$, or empty), the comment-enabled arm matches or exceeds the uncommented arm's constraint satisfaction
  in every group
  (\autoref{tab:outcomes}); only \OutcomeOverprunedShare{} of worlds end empty because the comment-enabled
  maintainer deleted a previously-learned requirement.

Finally, we show stronger maintainers ratchet harder and gain more from comments: across \CapacityTiers{}
tiers of one model family with a fixed executor, uncommented excess rises from \CapacityControlRange{}, and
at the top tier comments improve both size and constraint satisfaction (\autoref{tab:capacity}).

\subsection{Comments beat existing context maintainers on real prompts and code}
\label{sec:wild}

  On real-world and coding-convention suites, comment-enabled agentic maintainers construct the smallest prompts
  (\BeyondGapLow{} to \BeyondGapHigh{} of $|D_\star|$ below the ACE and Dynamic Cheatsheet maintainers) at
  constraint satisfaction no baseline
  beats (paired differences \BeyondSatLow{} to \BeyondSatHigh{}; \autoref{tab:beyond}).

  We invert WildIFEval \citep{lior2025wildifeval} and CodeIF \citep{yan2025codeif} the same way, carrying the
  test to real prompts and to coding conventions.
  WildIFEval's hidden constraints are real user requests --- what to include or avoid, format,
  tone, length; CodeIF's are a programming task's conventions --- names, signatures, control flow, imports.
  Neither suite is mechanically verifiable, so in both an arm-blind LLM judge scores each constraint and
  delivers failures to the maintainer, as review feedback does in agentic coding.

\input{assets/table_beyond.tex}
  We further add two context management baselines: ACE \citep{zhang2025ace} adds items, tracking helpful and
  harmful counts, while continually pruning duplicates; Dynamic Cheatsheet \citep{suzgun2025dynamiccheatsheet}
  curates the whole list each round, refining what feedback supports and dropping what it judges obsolete.
  Holding the executor, judge, worlds, feedback and round count ($T{=}\BeyondT{}$) fixed, comment-enabled
  maintainers shrink the prompt from \BeyondWildNoneExcess{} to \BeyondWildCommitExcess{} excess size on
  WildIFEval and from \BeyondCodeNoneExcess{} to \BeyondCodeCommitExcess{} on CodeIF, using fewer tokens than
  either baseline (\autoref{tab:beyond}).

\subsection{Comments improve instruction-following in real repositories}
\label{sec:live}

\input{assets/table_live.tex}

  On real repositories, comment-enabled agentic maintainers prune prompts that beat the repositories' actual rewrites
  at constraint satisfaction, including those written by human maintainers.

  Over \LiveJointPairs{} judged real task--instruction pairs, comment-enabled agentic pruning raises pair-weighted
  constraint satisfaction by \LiveJointPostGap{} over the post-rewrite file ($p = \LiveJointPostP$, paired sign-flip
  over repositories; \autoref{tab:live}), and by \LiveUnsignedPostGap{} ($p = \LiveUnsignedPostP$) where no agentic
  maintainer is identifiable behind the rewrite.
  Against uncommented agentic pruning, it deletes \LiveCommitDropped{} of a file (vs. \LiveControlDropped{}) yet
  wins constraint satisfaction by \LiveJointGap{} ($p = \LiveJointP$).
  \apxref{sec:appendix:live} details world selection, lookahead-free comment reconstruction, agentic
  pruning, validation task reconstruction, two-judge scoring and validation.

  Comments thus hint at an exit from the growth--rewrite--regrowth cycle of public repositories
  (\autoref{sec:rewrite}).
  Rewrites cut \LivePostShrink{} of a file but lose \LivePostSatLoss{} of constraint satisfaction; we surmise
  wrongly cut
  instructions return as failures and drive regrowth, and that comments limit cuts to instructions whose latent
  reasoning lapsed, shrinking prompts and improving constraint satisfaction.

Under assignment, then, prompt comments encoding latent reasoning yield the smallest prompt at matched
constraint satisfaction, outperform existing context maintainers and at small scale buy back constraint
satisfaction in
real repositories.

%% file: assets/table_access.tex
\begin{table}[t]
  \begin{captionedtable}{tab:access}{
    Agentic maintainers that write and read back their latent reasoning as comments reach the smallest size at
    matched constraint satisfaction. Top: the executor alone under IFEval's instructions $D_\star$ and their
    minimum verified cover $D_V$ (leave-one-out). Bottom: maintainers whose comments are absent (\textsc{None}),
    written but never read (\textsc{WriteOnly}), another task's (\textsc{ShuffledNoise}) or their own
    (\textsc{PromptComments}). Excess is prompt size beyond $D_V$ (lower is better); satisfaction is the share of
    hidden constraints met at the start ($t{=}0$) and over the last 3 of \AccessT{} rounds ($t{=}T$). \AccessWorlds{} worlds, \AccessSeeds{} seeds; $\pm$ 95\pct{} CI; \textbf{bold}: smaller than the next arm
    at matched satisfaction (paired, $p < 0.05$).
  }
  \centering\small
  \setlength{\tabcolsep}{4pt}
  \begin{tabularx}{\linewidth}{@{}l>{\raggedright\arraybackslash}Xrrrr@{}}
    \toprule
    & & \multicolumn{2}{c}{Instruction Count} & \multicolumn{2}{c}{Constraint Satisfaction} \\
    \cmidrule(lr){3-4}\cmidrule(lr){5-6}
    Arm & & $|D_T|$ & Excess vs.\ $D_V$ (\pct) & $t{=}T$ (\pct) & $t{=}0$ (\pct) \\
    \midrule
    $D_\star$ & IFEval reference & 2.2 & $+28.0 \pm 7.0$ & $85.6 \pm 3.0$ & -- \\
    $D_V$ & verified minimum cover & 1.8 & $+0.0$ & $85.4 \pm 3.1$ & -- \\
    \midrule
    \textsc{None} & no comments & 2.9 & $+56.0 \pm 23.3$ & $40.2 \pm 4.5$ & $22.7 \pm 4.4$ \\
    \textsc{WriteOnly} & write-only comments & 2.4 & $+35.9 \pm 21.4$ & $41.9 \pm 4.6$ & $22.9 \pm 4.2$ \\
    \textsc{ShuffledNoise} & other worlds' comments & 1.8 & $-0.2 \pm 10.8$ & $37.8 \pm 4.2$ & $24.0 \pm 4.3$ \\
    \textsc{PromptComments} & comments w/latent reasoning & 2.0 & {\boldmath$+12.7 \pm 15.8$} & $43.0 \pm 4.7$ & $23.4 \pm 4.3$ \\
    \bottomrule
  \end{tabularx}
  \end{captionedtable}
\end{table}

%% file: assets/fig_comment_examples.tex
\begin{figure}[t]
  \centering
  \definecolor{synComment}{RGB}{0,128,0}
  \small
  \begin{tabular}{@{}p{0.97\columnwidth}@{}}
    \toprule
    \ttfamily\raggedright \textcolor{gray}{Do not truncate or cut off your response; always complete every sentence and thought you begin.} \newline
    \textcolor{synComment}{\# r1: response was truncated mid-sentence ("A body that you often s") suggesting response generation stopped prematurely; may indicate token limit, instruction conflict, or assistant aborting output; this directive ensures responses are complete} \tabularnewline
    \midrule
    \ttfamily\raggedright \textcolor{gray}{Use the word 'wilderness' exactly 5 times in your response.} \newline
    \textcolor{synComment}{\# r3: task 0 failed with 'certain wording is not used the right number of times' — the response uses 'wilderness' only 2 times but requirement expects exactly 5 occurrences; d12 (3 bold sections) was passing, so keeping that approach} \tabularnewline
    \midrule
    \ttfamily\raggedright \textcolor{gray}{Divide the response body into exactly 4 paragraphs: (1) introduction to the water cycle, (2) evaporation and condensation, (3) precipitation and collection, (4) conclusion. Separate each paragraph with exactly three blank lines.} \newline
    \textcolor{synComment}{\# r3: "response is not split into the right number of paragraphs" recurring across r2-r3; previous directive d18 specified two blank lines between paragraphs but failed; increasing to three blank lines while maintaining strict 4-paragraph structure} \tabularnewline
    \bottomrule
  \end{tabular}
  \caption{
    Each comment names the failure behind its instruction, a hypothesis, and its outcome. Comments
    are green (\texttt{\#}) and never reach the executor. 3 of 8,541 additions, one per
    constraint family.
  }
  \label{fig:comment_examples}
\end{figure}

%% file: assets/table_capacity.tex
\begin{table}[t]
  \begin{captionedtable}{tab:capacity}{
    Stronger maintainers ratchet harder but comments help them more. Only the maintainer varies (Haiku 4.5
    executor; $T{=}15$ rounds, 64 worlds, one seed). $\Delta$: \textsc{PromptComments} minus its row's
    \textsc{None}, in points; $\pm$: half a 95\% bootstrap CI; \textbf{bold}: largest $\Delta$ whose CI excludes zero.
  }
  \centering\small
  \begin{tabularx}{\linewidth}{@{}>{\raggedright\arraybackslash}Xrrrr@{}}
    \toprule
    & \multicolumn{2}{c}{\textsc{None}} & \multicolumn{2}{c}{\textsc{PromptComments} vs.\ \textsc{None}} \\
    \cmidrule(lr){2-3} \cmidrule(lr){4-5}
    Maintainer & $|D_T|/|D_\star|$ & Sat.\ (\pct, $t{=}T$) & $\Delta$ Excess Size (pp) & $\Delta$ Sat.\ (pp) \\
    \midrule
    Haiku 4.5 & $1.68 \pm 0.58$ & $40.9 \pm 7.7$ & $-70.6 \pm 56.5$ & $-1.8 \pm 4.8$ \\
    Sonnet 5 & $3.07 \pm 0.47$ & $38.5 \pm 7.6$ & $-189.3 \pm 44.4$ & $+2.9 \pm 6.4$ \\
    Opus 5 & $6.72 \pm 0.88$ & $45.3 \pm 8.7$ & {\boldmath$-351.3 \pm 70.9$} & {\boldmath$+8.3 \pm 7.7$} \\
    \bottomrule
  \end{tabularx}
  \end{captionedtable}
\end{table}

%% file: assets/table_beyond.tex
\begin{table}[t]
  \begin{captionedtable}{tab:beyond}{
    Comments keep the prompt smallest beyond format constraints, against the uncommented maintainer and ACE-style
    and Dynamic Cheatsheet maintainers adapted to binding directives. Same executor, judge, worlds, feedback and \BeyondT{} rounds;
    only the update rule differs. Excess size $|D_T|/|D_\star|{-}1$, satisfaction over the last 3 of \BeyondT{}
    rounds ($t{=}T$), maintainer tokens per round. \BeyondWorlds{} worlds per suite, one seed; \textbf{bold}: best arm whose paired 95\% interval against
    the runner-up excludes zero.
  }
  \centering\small
  \begin{tabularx}{\linewidth}{@{}l>{\raggedright\arraybackslash}Xrrr@{}}
    \toprule
    Suite & Maintainer & Excess size (\pct) & Sat.\ (\pct, $t{=}T$) & Tokens/round \\
    \midrule
    CodeIF & \textsc{None} & $+1.7$ & $34.4$ & 2.9K \\
     & ACE-style & $+29.9$ & $28.3$ & 10.9K \\
     & Dynamic Cheatsheet & $+101.0$ & $33.2$ & 8.3K \\
     & \textsc{PromptComments} & {\boldmath$-32.7$} & $35.7$ & 5.7K \\
    \midrule
    WildIFEval & \textsc{None} & $+38.5$ & $42.7$ & 2.4K \\
     & ACE-style & $+95.7$ & $40.9$ & 8.7K \\
     & Dynamic Cheatsheet & $+305.7$ & $48.8$ & 6.2K \\
     & \textsc{PromptComments} & {\boldmath$+14.5$} & $47.8$ & 4.7K \\
    \bottomrule
  \end{tabularx}
  \end{captionedtable}
\end{table}

%% file: assets/table_live.tex
\begin{table}[t]
  \begin{captionedtable}{tab:live}{
    Reconstructed comments let an agentic maintainer prune real context files more boldly and to higher satisfaction.
    \LiveJointPairs{} task--instruction pairs over \LiveFiles{} worlds, scored on every arm by two reasoning judges;
    one seed. Instructions: mean $|D|$ per world; satisfaction: pair-weighted; \textbf{bold}: column best above \textsc{None} (paired sign-flip over worlds, $p < 0.05$).
  }
  \centering\small
  \begin{tabular*}{\linewidth}{@{}ll@{\extracolsep{\fill}}r@{\extracolsep{0pt}\hspace{2\tabcolsep}}r@{}}
    \toprule
    Arm & Maintainer & Instructions & Satisfaction (\pct) \\
    \midrule
    $D_{\text{pre}}$ & repository maintainers (pre-rewrite) & $111.4$ & $42.1$ \\
    \midrule
    $D_{\text{post}}$ & repository maintainers (post-rewrite) & {\boldmath$27.4$} & $37.5$ \\
    \textsc{None} & agentic maintainer (no comments) & $104.7$ & $37.6$ \\
    \textsc{PromptComments} & agentic maintainer (with comments) & $92.4$ & {\boldmath$44.2$} \\
    \bottomrule
  \end{tabular*}
  \end{captionedtable}
\end{table}

%% file: sections/05-related.tex
\section{Related Work}
\label{sec:related}

\paragraph{Empirical studies of agentic prompts.}
Prior measurement establishes that agentic prompts earn their keep and only grow.
\citet{chatlatanagulchai2025agentreadmes} characterize \texttt{AGENTS.md}-class files, which
accumulate content and rarely shed it; \citet{tafreshipour2025promptinthewild} track prompt evolution in
repositories, where additions are the most frequent edit type; and \citet{lulla2026contextfiles}
associate them with faster agent runs and fewer output tokens. We measure that growth over the lifetimes of individual
instructions, fine enough to estimate a deletion hazard and test which mechanism it fits
(\autoref{sec:hazard}).

\paragraph{Agentic memory and context management.}
Agent memory systems forget against an observable staleness proxy: MemGPT evicts on context overflow
\citep{packer2023memgpt}; Mem0 deletes on detected contradiction, and its graph variant and Zep invalidate
superseded entries by timestamp \citep{chhikara2025mem0, rasmussen2025zep}; FSFM catalogs mechanisms from passive
decay to safety-triggered deletion \citep{fsfm2026selective}. Authored instructions do not have a similar
proxy, as neither age nor disuse make it obsolete; instead, we show its latent reasoning and its subsequent
correctness drives whether it should remain (\autoref{sec:theory}, \autoref{sec:hazard}).

  ACE \citep{zhang2025ace} and Dynamic Cheatsheet \citep{suzgun2025dynamiccheatsheet} maintain context from execution
  feedback: ACE appends itemized lessons, pruning only duplicates or at a cap; Dynamic Cheatsheet re-curates
  the list each round, dropping stale ones. Both curate advisory knowledge filtered at inference, so
  excess is cheap; agentic instructions bind every output, and instruction-following degrades rapidly as they
  accumulate. Neither records why entries exist, so neither separates needed instructions from surplus;
  both yield larger agentic prompts than comments (\autoref{tab:beyond}).

\paragraph{Organizational behavior.}
Organizational written rules are artifacts people maintain for decades, and are well studied.
Lehman's laws attribute a program's growth to a changing environment \citep{lehman1980laws}, the demand-side
account; \citet{zhou1993dynamics} treats rule change as a stochastic process whose rate depends on a rule's
own age; \citet{schulz1998limits} shows rule birth rates falling as rules densify; and
\citet{march2000dynamics} assembles both into an account of rule births, revisions, and suspensions. We adopt
their event-history form and add the factor it lacks, latent reasoning's recoverability
at age $a$, $\bar \rho(a)$, separated from compositional and staleness effects in agentic prompts
(\autoref{eq:hazard}, \autoref{sec:hazard}).

\paragraph{Software engineering best practices.}
Software engineering solved comparable challenges decades ago via syntactic structures and engineering
conventions. Literate programming argued that a program is addressed to a human reader \citep{knuth1984literate};
\citet{mcconnell2004codecomplete} makes commenting the \emph{why} standard practice;
\citet{latoza2006mentalmodels} find recovering that latent reasoning is developers' most serious problem;
architecture decision records institutionalize recording it at decision time \citep{nygard2011adr}; and
commit messages exist to carry it and routinely fail \citep{tian2022commitmessages}. We carry both the
comment and what it should record into agentic prompts, and show that comments
encoding an instruction's latent reasoning cut excess size at parity constraint satisfaction (\autoref{sec:protocol}).

However, code comments are notoriously unfaithful, and entire lines of research study how they diverge
from code. Comments and the code they describe co-evolve poorly \citep{fluri2007coevolution}; detecting and
addressing comment--code inconsistency are their own ML-for-SWE tasks \citep{panthaplackel2020comments,
panthaplackel2021jit}. Our ablations show the same failure in agentic prompts: low-information,
poorly structured or misleading latent reasoning forfeits the gain (\autoref{sec:protocol},
\autoref{tab:ablations}).

\paragraph{Continual learning.}
Continual learning is our closest cousin but primarily targets tasks with drifting objectives. Networks
trained on a task sequence overwrite what earlier tasks taught them \citep{mccloskey1989catastrophic,
french1999catastrophic}; rehearsal and regularization \citep{kirkpatrick2017overcoming} help preserve what
shifting objective might otherwise destroy; L2P, DualPrompt, and CODA-Prompt move that burden into the input,
learning \emph{fixed-size} prompt pools from which frozen backbones selects, binding growth by construction
\citep{wang2022l2p, wang2022dualprompt, smith2023coda}.
Although our task is stationary, we borrow continual learning frameworks because a fixed resource binds in
both: shifting objectives overwrite fixed parameters there (catastrophic forgetting), whereas here intermittent failures
fill an instruction budget capped by instruction-following drop-offs \citep{jaroslawicz2025ifscale}
(catastrophic remembering; \autoref{eq:ratchet}, \autoref{sec:protocol}).

\paragraph{Instruction following.}
IFEval scores responses against verifiable constraints \citep{zhou2023ifeval}; FollowBench grades
progressively added constraint levels and WildIFEval collects real requests carrying many at once, both
documenting that following degrades as the required set grows
\citep{jiang2024followbench, lior2025wildifeval}.
Those benchmarks price the constraints a response must satisfy, the harm our motivation rests on. We invert
them to make the minimum cover observable, without which excess size is unmeasurable (\autoref{sec:testbed},
\autoref{sec:wild}).

%% file: sections/06-discussion.tex
\section{Discussion}
\label{sec:discussion}
If English is the new code, why don't we have comments yet?

We show two things: (i) how agentic prompts grow is consistent with an instruction's latent reasoning
decaying faster than the instruction, and with no rival we test (\autoref{sec:hazard}), and (ii) under
censored feedback and a maintainer-only comment channel, saving that latent reasoning at write time reduces
instruction count and wins back constraint satisfaction (\autoref{sec:protocol}, \autoref{sec:wild},
\autoref{sec:live}).

Neither half should surprise us. Software engineering identified and solved the same problem decades ago and continual
learning is working on solving its dual --- catastrophic forgetting --- right now. Both differ from
ours in detail but not in principle: whatever licenses an update must outlive it. LLM agents began driving
real work before either lesson reached them, so we borrow from both now.

Our results open three lines of work. Coding agent developers can give prompts a comment syntax, so
an instruction's latent reasoning can reach the next maintainer; our prototype, while promising, is just a
prototype. Maintainers who manage agentic coding prompts can identify best practices, starting from the software
engineering ones that, like the comment, already carry over. Finally,
researchers can look for methods that beat comments, since prompt maintenance is continual learning over text
and nobody has built its analogue of rehearsal or regularization, and for benchmarks that measure it at
representative constraint counts and for constraints not mechanically verifiable.

Every field is borrowing from AI right now. We should borrow back just as readily. The answer to
catastrophic remembering was forty years old and one field over, and the next one might be too.

%% file: sections/08-llm-usage.tex
\section*{AI Use Statement}
We use LLMs liberally as a delegated ``compiler'': we supply the theoretical model, experimental framework,
falsifiable hypotheses, final judgments, coding and experimental standards, etc. and ask LLMs to carry out
the execution under the constraints we specify.

Specifically, we used generative AI tools to formulate mathematical claims (e.g. formalizing derivation of
\autoref{sec:theory} whose dynamics we outlined in prose), provide feedback on experiment design (e.g. reviewing our
experiment designs), and implement methods (e.g. the experiment and analysis code). To select the
prompt comment protocol (\apxref{sec:appendix:ladder}), we used an auto-research pipeline that autonomously and iteratively analyzed data,
interpreted results and proposed and refined hypotheses given a seed set of initial hypotheses. Our methods
also use LLMs as declared in-method tools: one summarizes each Inverse-IFEval world's instructions as its task objective
(\autoref{sec:testbed}), and an arm-blind LLM judge scores constraint satisfaction on WildIFEval
and CodeIF (\autoref{sec:wild}).

We have not used generative AI tools to generate synthetic data sets, help develop theoretical models or
conceptual frameworks or clean and reformat dataset.

Additionally, we used generative AI tools to create or modify scientific figures or images (e.g. the
plotting scripts behind each figure), create or edit software code (e.g. the scripts that build the
paper and generate its numbers), draft parts of a research paper (e.g. first drafts of sections and
appendix text, which we then rewrote), summarize or analyse existing literature (e.g. for
\autoref{sec:related}), edit a research paper to improve readability (e.g. tightening prose), identify
relevant literature (e.g. surfacing candidate related work), format references (e.g. BibTeX entries),
and propose a title or keywords for a research paper (e.g. candidate titles).

We have validated all AI-assisted work. We hand-checked all mathematical claims, rewrote all
main-body text, and code-reviewed all experiment and analysis code. A generator script emits every table,
figure and reported number from the canonical experimental artifacts. Outside the declared in-method uses
above, no generative tool generates or manipulates data. Finally, we screened the manuscript with Pangram for
inadvertent plagiarism.

We take responsibility for the final content of this work, including text, claims or artifacts produced
with the aid of generative AI.

%% file: sections/09a-ethics.tex
\section*{Ethics Statement}
We mine public repository histories, reduce commit authorship to a count of distinct editors, and
release no identities or repository content. Our recommendation carries the main risk: automated
deletion from latent reasoning will remove instructions whose latent reasoning was real. Deployments should
keep a person in the deletion path and exclude safety-relevant instructions.

%% file: sections/09b-reproducibility.tex
\section*{Reproducibility Statement}

\input{assets/table_packages.tex}
\autoref{tab:packages} pins every package a reported number passes through to the canonical run's
\texttt{env.lock}. The lower block lists our own estimators (\apxref{sec:appendix:hazard}) and
the \LabVerifierTypes{} reimplemented verifiers, which return censored complaints against a
maintained prompt.

We train no model. The corpus pipeline runs on CPU containers (8 vCPU, 16\,GiB) over \FieldRepos{}
repositories. The controlled experiment's canonical run cost \CostCalls{} model calls,
\CostTokens{} tokens and \CostModelHours{} model-hours across roles, seeds and arms. Call counts
match across arms by construction; only maintainer input tokens differ
(\CostMaintainerInRange{}). We pin model versions, decode at API defaults, and ran no
hyperparameter search beyond selecting the comment protocol (\apxref{sec:appendix:ladder}).

The sampling frame is the repository list of \citet{chatlatanagulchai2025agentreadmes}, which
carries no license, so we use only its URLs and re-derive all content from public histories.
IFEval \citep{zhou2023ifeval} is Apache~2.0. We redistribute no repository content and release
only derived per-instruction tables and code.

The corpus is \FieldRepos{} repositories, \FieldMultiVersionFiles{} multi-version files,
\FieldMatches{} tracked transitions and \FieldSpells{} spells, \FieldDeletions{} ending in
deletion (\autoref{tab:funnel}).
\FloatBarrier

%% file: assets/table_packages.tex
\begin{table}[htb]
  \begin{captionedtable}{tab:packages}{
    The packages and models each reported number passes through. Top: packages at the versions frozen in the
    canonical runs' \texttt{env.lock}, identical across runs. Middle: models by pinned API snapshot and serving
    route; \texttt{claude-opus-5-5} reasons at effort high, and OpenRouter serves \texttt{gemini-3.8-flash} from
    Google AI Studio only (no fallback) at effort medium. Bottom: our own implementations
    (\autoref{sec:appendix:hazard}, \autoref{tab:frailty}). Satisfaction rates are against our
    verifiers, not published IFEval numbers.
  }
  \centering\small
  \begin{tabularx}{\linewidth}{@{}ll>{\raggedright\arraybackslash}Xl@{}}
    \toprule
    Package or model & Version or route & Determines & Where \\
    \midrule
    \texttt{rapidfuzz} & 3.14.5 & cross-version match at similarity $\ge$ \MatchFuzzyThreshold{}: every spell, death and hazard number & \autoref{sec:appendix:construction} \\
    \texttt{lifelines} & 0.30.3 & Nelson--Aalen slope $\FieldSlopeCommit$ & \autoref{sec:hazard} \\
    \texttt{scipy} & 1.17.1 & Poisson likelihood of every hazard family; effort-dilution regression & \autoref{tab:hazardforms} \\
    \texttt{numpy} & 2.4.6 & repository bootstrap draws (\HazardBoot{}, seed 0) behind every field interval & \autoref{sec:appendix:hazard} \\
    \texttt{pandas} & 2.3.3 & every aggregate, ratio and risk table & -- \\
    \texttt{anthropic} & 0.120.2 & every \texttt{claude-*} call below & \autoref{sec:lab} \\
    \texttt{openai} & 2.53.0 & \texttt{gpt-5.6-luna} (OpenAI API) and \texttt{gemini-3.8-flash} (OpenRouter) judge calls & \autoref{sec:appendix:live} \\
    \midrule
    \texttt{claude-haiku-4-5} & Anthropic & executor everywhere; maintainer in \autoref{sec:testbed}--\ref{sec:protocol}; WildIFEval and CodeIF judge & \autoref{sec:appendix:testbed} \\
    \texttt{claude-sonnet-5} & Anthropic & objectives $o_j$; maintainer in \autoref{sec:wild}--\ref{sec:live} & \autoref{sec:appendix:testbed} \\
    \texttt{claude-opus-5} & Anthropic & top capacity-tier maintainer & \autoref{tab:capacity} \\
    \texttt{claude-opus-5-5} & Anthropic & comments reconstructed from repository history & \autoref{sec:appendix:live} \\
    \texttt{gpt-5.6-luna} & OpenAI & live-repository judge & \autoref{sec:appendix:live} \\
    \texttt{gemini-3.8-flash} & OpenRouter & live-repository judge & \autoref{sec:appendix:live} \\
    \midrule
    piecewise-exponential EM & ours & gamma frailty slopes and absorbed share & \autoref{tab:frailty} \\
    DerSimonian--Laird & ours & two-stage combination of decay, frailty and forms & \autoref{tab:hazardforms} \\
    penalized IRLS & ours & file-fixed-effect spline; \FieldFEFiles{} files & \autoref{sec:appendix:hazard} \\
    \LabVerifierTypes{} verifiers & ours & every satisfaction rate in \autoref{sec:lab} & \autoref{tab:verifiers} \\
    \bottomrule
  \end{tabularx}
  \end{captionedtable}
\end{table}

%% file: sections/07-limitations.tex
\section*{Limitations}
\autoref{sec:field} is observational: it characterizes growth and rejects the rivals we test,
while \autoref{sec:lab}, where comments are assigned, carries the causal claim. The declining hazard survives sweeps
of the measurement thresholds (\autoref{tab:thresholds}) and the file's size at each commit
(\apxref{sec:appendix:dilution}); residual matcher error biases the age slope toward zero, against our finding
(\apxref{sec:appendix:tracking}).

\autoref{sec:protocol} isolates the mechanism in small, stationary worlds: covers of two or three instructions,
\HorizonT{} rounds, one model family and dense, verifiable feedback. Task drift, rarely triggered instructions and
executor upgrades lie outside them, and rewrites arise there only unprompted (\apxref{sec:appendix:worldrewrites}).
Dense feedback and a stationary task let trial and error stand in for comments, so our size gains are a lower
bound.

Comments reach the minimal size, not $D_\star$'s constraint satisfaction (\CoverDstarSat{}): every maintainer arm ends between
\AccessNoiseSat{} and \AccessCommitSat{} (\autoref{tab:access}), which constitutes a ceiling on what current agentic
maintainers can learn from censored feedback in our environment.

\autoref{tab:access} and its downstream results in Sections~\ref{sec:testbed}--\ref{sec:protocol} run
\AccessSeeds{} seeds; the \HorizonT{}-round headline, capacity tiers, comment ablations, coding-convention
suites and live-repository prune run one.
The live-repository prune is also a single maintenance step, with comments reconstructed from history rather than
written at addition per protocol, which likely reduces measurable impact.
LLM judges, not code verifiers, score WildIFEval, CodeIF and the live tasks.

Our scope is agentic prompts on public GitHub (\FieldRepos{} repositories); no result speaks to non-English
instructions, system prompts or agent skill files.

%% file: sections/10a-appendix-observational.tex
\section{The Observational Corpus}
\label{sec:appendix:corpus}

\subsection{Instruction lifetimes from commit history}
\label{sec:appendix:construction}
\label{sec:appendix:matching}
\label{sec:appendix:tracking}

\input{assets/table_funnel.tex}
We sample the public GitHub repositories of \citet{chatlatanagulchai2025agentreadmes} that hold a
\texttt{CLAUDE.md}, \texttt{AGENTS.md} or \texttt{copilot-instructions.md} (\autoref{tab:funnel}), walking every
version of every agentic prompt forward so that no stage conditions on survival. An \emph{instruction} is one
clause-level imperative, split on markdown structure, then sentences; headings, code, tables, bold labels and
fragments under \SegMinWords{} words are \emph{payload}, counted in \autoref{sec:growth} but outside the risk set.
Every ratio we report is invariant to rescaling $|D_t|$.

\input{assets/table_gates.tex}
A three-stage cascade (exact, normalized, then fuzzy above \MatchFuzzyThreshold{}) matches instructions across
versions; an unmatched old instruction dies, an unmatched new one is born. The matcher scores
\GateBPrecision{} precision and \GateBRecall{} recall, each above its \GateBFloor{} floor (\autoref{tab:gates}). Missed
rewordings add hazard to old instructions, so the decline we report is a lower bound.

\subsection{Files grow in count, not length}
\label{sec:appendix:decomposition}

\input{assets/table_complexity.tex}
Files add rules rather than lengthen them (\autoref{tab:complexity}), and the repository's total prompt text grows
\FieldGrowthPayloadPct{}, so the count is not text migrating between files. The 90th-percentile file ends its
history at \FieldOpNinety{} instructions. Of \FieldMultiVersionRepos{} multi-version repositories, \FieldFracGrow{}
gain instructions and \FieldFracShrink{} lose them;
\autoref{sec:growth}'s net \FieldNetPerCommit{} per commit pools \FieldTouchingCommits{} commits.

\subsection{Rewrites censor deaths and answer growth}
\label{sec:appendix:censoring}

\input{assets/table_rewrite.tex}
Censored with migrations as competing risks (\autoref{sec:measurement}; floor \RewriteMinDirectives{}
instructions), rewrite deaths outnumber hazard deletions \FieldRewriteToDeletionRatio{} to one: maintainers
replace rather than prune, as \autoref{eq:ratchet} predicts. Rewrites answer growth, not failure:
\ObsLengthSubjects{} of \ObsRewriteCommits{} rewrite-commit subjects say the file grew too long or needed trimming;
none says it stopped working. Growth after a rewrite is no slower (\autoref{tab:rewrite}), so the ratchet is
not self-limiting.

\subsection{Hazard families and rival mechanisms}
\label{sec:appendix:hazard}
\label{sec:appendix:frailty}
\label{sec:appendix:dilution}

Every estimator reads deletion risk against age in commits touching the file (\autoref{eq:hazard}), censoring
competing risks and end of history, over the \autoref{tab:funnel} hazard population (histories of
${\ge}\HazardMinDuration$ commits). Three estimators agree on sign and shape: a Nelson--Aalen estimate smoothed at
\HazardBandwidthCommits{} commits over ages $0$ to \HazardAgeMax{}, whose slope \autoref{sec:hazard} quotes; a
penalized spline with file fixed effects (\FieldFEFiles{} files); and the parametric fits below. Intervals
bootstrap over repositories (\HazardBoot{} draws).

\input{assets/table_hazard_forms.tex}
Each form in \autoref{tab:hazardforms} decides whether \autoref{eq:size} converges. Per-repository fits pool by
random effects, so between-repository spread enters $\tau^2$, not curvature favouring a bounded family.
The geometric fit also estimates an excess probability $s = \HazardGeoExcess$.

\input{assets/table_frailty.tex}
Heterogeneity alone could produce the decline: a gamma frailty with variance $\theta$ gives
$h_{\text{pop}}(a) = h_0(a)/(1 + \theta H_0(a))$, which declines even from a flat $h_0$; \autoref{tab:frailty}
gives the share of slope it absorbs.

Fixed effort dilution fails: Poisson regression of deletions on instructions at risk gives exponent
$\gamma = \FlowGammaDeletionCI$, against $0$ under fixed effort and $1$ under
per-instruction risk; additions give $\FlowGammaAdditionCI$. Partial dilution may
contribute but does not carry the age effect: controlling for file size at each commit, the age slope stays at
$\RecallAgeSlopeCI$ per commit (\RecallRepos{} repositories, repository-clustered), negative in
\RecallBandsNegative{} of \RecallBands{} size bands.

Lost relevance does not depend on who commits, yet an instruction's own author deletes it $\RecallOwnAuthor\times$
as often as other committers ($\RecallOwnAuthorCI$), and $\RecallOwnSection\times$ ($\RecallOwnSectionCI$) as often
as their neighbors' instructions in the same commit and section (\RecallSectionStrata{} commit-sections).

\subsection{Recorded latent reasoning and relearning}
\label{sec:appendix:rationale}

Real instructions rarely record their latent reasoning, as \autoref{sec:lab}'s uncommented arms assume:
\ObsReasonPhrase{} \ObsReasonPhraseCI{} of \FieldSpells{} instructions contain a reason phrase
(because, so that, \ldots), and hand-coding \ObsFailureCoded{} keyword matches puts the share stating the failure
they fix at \ObsFailureStated{} \ObsFailureStatedCI{}. Files that record reasons are rewritten less: net of size and
age, a file ever holding a reason phrase has $\ObsRationaleRewriteHR\times$ the rewrite hazard
($\ObsRationaleRewriteHRCI$, \ObsRewriteEvents{} rewrites, repository-clustered).

Files then relearn what rewrites cut: \ObsReaddRewrite{} \ObsReaddRewriteCI{} of additions restating a former
instruction restore one a rewrite removed. Across \ObsAdditions{} additions, each doubling of a file's version
index lowers the share resembling nothing the file held by \ObsSlopeNovel{} ($\ObsSlopeNovelCI$) and raises the
restating share by \ObsSlopeRelearned{} ($\ObsSlopeRelearnedCI$); size-matched donor files stay flat.
Forgetting and relearning predict this decline; relevance drift predicts the reverse. Every analysis here is
keyword- or similarity-coded.

\subsection{Sensitivity to the measurement thresholds}
\label{sec:appendix:sweep}

\input{assets/table_thresholds.tex}
The age slope is negative, with an interval excluding zero, in all \SweepCells{} cells of
\autoref{tab:thresholds}, and does not track how much the rule censors. Geometric is the best form in
\SweepFormHeldCells{} cells, by at least \SweepFormMarginMin{} in $R^2_w$, with decay within
\SweepGeoDecayRange{}. Post-rewrite growth outpaces pre-rewrite growth in \SweepRegrowthFasterCells{} of
\SweepRewriteCells{} rewrite-fraction cells, failing at the fraction censoring the most deaths.

%% file: assets/table_funnel.tex
\begin{table}[H]
  \begin{captionedtable}{tab:funnel}{
    What each pipeline stage keeps, from sampling frame to estimation panels. \emph{Repos.}:
    repositories represented; $N$: count in the stage's unit. The death rows partition ended spells
    under the canonical rewrite rule (\autoref{sec:appendix:censoring}); the two-stage row is the
    repositories with at least five deletions (\autoref{tab:hazardforms}); the collapsed panel is
    \autoref{fig:hazard}'s cut.
  }
  \centering\small
  \begin{adjustbox}{max width=\linewidth}
  \begin{tabular}{@{}llrr@{}}
    \toprule
    Stage & Rule & Repos. & $N$ \\
    \midrule
    repositories & $\ge 1$ tracked context file & 1,867 & 1,867 \\
    multi-version files & $\ge 2$ versions & 1,558 & 1,801 \\
    tracked transitions & consecutive version pairs & 1,510 & 299,440 \\
    instruction spells & matched across transitions & 1,856 & 247,694 \\
    deaths: deletion & the hazard's event & 1,124 & 40,355 \\
    deaths: rewrite-censored & $\ge$ 75\% of live instructions killed, floor 5 & 640 & 84,233 \\
    deaths: migration-censored & text reappears in a sibling file & 37 & 682 \\
    hazard population & file history $\ge$ 3 commits & 1,095 & 177,995 \\
    two-stage fits & $\ge 5$ deletions per repository & 430 & -- \\
    collapsed panel & converges under every family, $\le 4$ half-lives & 284 & 15,574 \\
    \bottomrule
  \end{tabular}
  \end{adjustbox}
  \end{captionedtable}
\end{table}

%% file: assets/table_gates.tex
\begin{table}[H]
  \begin{captionedtable}{tab:gates}{
    Every measurement gate passes on the canonical corpus. Columns: the gate's threshold and this
    run's value. Gate~B is measured on a 50-transition hand-annotated pool.
    \emph{Grow:shrink}: share of multi-version repositories gaining instructions over share losing
    them; \emph{tail share}: fraction of deletions at or above the median instruction age.
  }
  \centering\small
  \begin{tabularx}{\linewidth}{@{}c>{\raggedright\arraybackslash}Xrrc@{}}
    \toprule
    Gate & Criterion & Threshold & Observed & Pass \\
    \midrule
    0 & Median per-repo net $\Delta D > 0$ & $> 0$ & $+7$ & $\checkmark$ \\
    0 & Grow:shrink ratio & $> 1.5$ & 2.42 & $\checkmark$ \\
    A & Deletions excluding rewrites & $\ge 300$ & 28,426 & $\checkmark$ \\
    A & Tail share at/above median age & $\ge 20.0\%$ & 57.0\% & $\checkmark$ \\
    B & Matcher precision / recall & $\ge 0.85$ & 1.000 / 0.933 & $\checkmark$ \\
    -- & Median rel.\ growth, all multi-version files: count vs.\ length & count $>$ length & $1.193$ vs.\ $1.028$ & $\checkmark$ \\
    \bottomrule
  \end{tabularx}
  \end{captionedtable}
\end{table}

%% file: assets/table_complexity.tex
\begin{table}[H]
  \begin{captionedtable}{tab:complexity}{
    Instructions grow far more numerous than longer. Mean percent change from each file's first
    version at fixed shares of its lifetime, $\pm$ 95\% CI half-width, over \FieldGrowthLenFiles{}
    files; \emph{Peak} is each curve's maximum over the lifetime grid. Count is evaluated stepwise
    between versions (\autoref{fig:hero:growth}), length linearly interpolated. Payload grows
    \FieldGrowthPayloadPct{}.
  }
  \centering\small
  \setlength{\tabcolsep}{4pt}
  \begin{tabular*}{\linewidth}{@{\extracolsep{\fill}}lrrrrr@{}}
    \toprule
    & \multicolumn{4}{c}{Share of file lifetime} & \\
    \cmidrule(lr){2-5}
    Change (\%) & 25\% & 50\% & 75\% & 100\% & Peak \\
    \midrule
    Count ($D_t$) & $+102.5 \pm 31.3$ & $+179.3 \pm 39.9$ & $+205.5 \pm 42.5$ & $+221.5 \pm 47.3$ & $+226.3 \pm 45.2$ \\
    Mean length ($\bar\ell_t$) & $+4.3 \pm 1.6$ & $+6.9 \pm 2.0$ & $+9.9 \pm 2.3$ & $+5.0 \pm 3.2$ & $+10.4 \pm 2.4$ \\
    \bottomrule
  \end{tabular*}
  \end{captionedtable}
\end{table}

%% file: assets/table_rewrite.tex
\begin{table}[H]
  \begin{captionedtable}{tab:rewrite}{
    The rewrite is a mass deletion, its shape does not depend on the window, and growth after it is
    no slower. \textbf{(a)} What rewrite commits do, corpus-wide. \textbf{(b)} The event study of
    \autoref{fig:massrewrite:event} at each window half-width $w$: median $D_t$ at $-w$, $0$ and
    $+w$ relative to the pre-rewrite version, over files observed at every offset, emptying events
    excluded; $\star$ marks the plotted width. \textbf{(c)} Growth within inter-rewrite segments
    ($\ge 2$ commits); segment $k$ starts at the $k$-th rewrite. The rewrite threshold is fixed at
    \RewriteFracThreshold{}.
  }
  \centering\small
  \begin{tabularx}{\linewidth}{@{}>{\raggedright\arraybackslash}Xrrrr@{}}
    \toprule
    \multicolumn{5}{@{}l}{\textbf{(a)} Anatomy of a rewrite commit} \\
    \midrule
    Rewrite commits & \multicolumn{4}{r@{}}{1,077} \\
    Median instructions killed / born & \multicolumn{4}{r@{}}{52 / 12} \\
    Events leaving the file empty & \multicolumn{4}{r@{}}{30.6\%} \\
    Events replacing $\ge$ half the dead text & \multicolumn{4}{r@{}}{33.0\%} \\
    Events carrying any cross-file migration & \multicolumn{4}{r@{}}{3.3\%} \\
    \midrule
    \multicolumn{5}{@{}l}{\textbf{(b)} Event study vs.\ window half-width} \\
    \midrule
    Window $w$ & $n$ & $k{=}{-}w$ & $k{=}0$ & $k{=}{+}w$ \\
    $\pm2$ & 252 & $-2.3\%$ & $-59.7\%$ & $-47.8\%$ \\
    $\pm3$ & 175 & $-6.9\%$ & $-63.6\%$ & $-51.3\%$ \\
    $\pm4$ & 134 & $-9.0\%$ & $-65.4\%$ & $-53.1\%$ \\
    $\pm5$ & 111 & $-11.2\%$ & $-70.3\%$ & $-52.4\%$ \\
    $\pm10^{\star}$ & 52 & $-21.3\%$ & $-73.1\%$ & $-60.7\%$ \\
    \midrule
    \multicolumn{5}{@{}l}{\textbf{(c)} Growth within inter-rewrite segments} \\
    \midrule
    Segment & $n$ & Median & Growing & Net $\Delta D$ \\
    0 & 1,641 & $\times1.35$ & 77.9\% & $+34.9$ \\
    1 & 454 & $\times1.40$ & 81.5\% & $+37.2$ \\
    2 & 119 & $\times1.45$ & 80.7\% & $+40.2$ \\
    3+ & 126 & $\times1.68$ & 85.7\% & $+52.3$ \\
    \bottomrule
  \end{tabularx}
  \end{captionedtable}
\end{table}

%% file: assets/table_hazard_forms.tex
\begin{table}[H]
  \begin{captionedtable}{tab:hazardforms}{
    The deletion hazard is geometric under both estimators; every bounded family fits worse. Each
    row fits $h(a)$ per repository by Poisson likelihood; \emph{Repos.}: fits converging with finite
    variance. Rows rank by $R^2_w$, fit to repository-weighted per-age rates (${<}0$: worse than
    their mean), as likelihoods sum over different repositories. \emph{Bounded iff}: finite expected lifetime;
    \emph{Decay}: its parameter (95\% CI); $\tau^2$: between-repository variance of log decay.
    Two-stage combines by DerSimonian--Laird over repositories with ${\ge}5$ deletions;
    hierarchical by maximum likelihood, no floor.
  }
  \centering\small
  \begin{adjustbox}{max width=\linewidth}
  \begin{tabular}{@{}llllrrrrl@{}}
    \toprule
    Estimator & Family & $h(a)$ & Bounded iff & Repos. & $\tau^2$ & Decay & 95\% CI & $R^2_w$ \\
    \midrule
    {\boldmath\textbf{two-stage (canonical)}} & {\boldmath\textbf{geometric}} & {\boldmath\textbf{$s\,e^{-ba}$}} & {\boldmath\textbf{never}} & {\boldmath\textbf{430}} & {\boldmath\textbf{0.97}} & {\boldmath\textbf{$b$ = 0.222}} & {\boldmath\textbf{$[0.199, 0.247]$}} & {\boldmath\textbf{0.815}} \\
     & Weibull & $k\,a^{p-1}$ & always & 439 & 0.22 & $p$ = 0.809 & $[0.762, 0.858]$ & 0.291 \\
     & constant & $s$ & always & 646 & 0.74 & -- & $[0.028, 0.032]$ & -0.313 \\
     & gamma mixture & $s/(1 + \theta s a)$ & $\theta < 1$ & 423 & 0.71 & $\theta$ = 7.034 & $[6.388, 7.747]$ & -0.482 \\
     & Burr XII & $\alpha s (as)^{c-1}/(1 + (as)^c)$ & $\alpha > 1$ & 448 & 0.76 & $\alpha$ = 0.151 & $[0.135, 0.168]$ & -0.739 \\
    \midrule
    hierarchical & geometric & $s\,e^{-ba}$ & never & 517 & 0.94 & $b$ = 0.241 & $[0.218, 0.267]$ & 0.658 \\
     & gamma mixture & $s/(1 + \theta s a)$ & $\theta < 1$ & 551 & 3.27 & $\theta$ = 8.351 & $[7.051, 9.891]$ & 0.543 \\
     & Weibull & $k\,a^{p-1}$ & always & 551 & 0.23 & $p$ = 0.813 & $[0.767, 0.863]$ & 0.082 \\
     & constant & $s$ & always & 921 & 1.28 & -- & $[0.022, 0.026]$ & -0.730 \\
     & Burr XII & $\alpha s (as)^{c-1}/(1 + (as)^c)$ & $\alpha > 1$ & 595 & 1.52 & $\alpha$ = 0.107 & $[0.094, 0.122]$ & -1.489 \\
    \bottomrule
  \end{tabular}
  \end{adjustbox}
  \end{captionedtable}
\end{table}

%% file: assets/table_frailty.tex
\begin{table}[H]
  \begin{captionedtable}{tab:frailty}{
    Within-repository frailty shrinks the age slope without reversing it. Both rows fit a
    piecewise-exponential hazard per repository (at least five deletions and two files;
    \S\ref{sec:appendix:frailty}), the second with a gamma frailty shared across the repository's
    files, and combine slopes by DerSimonian--Laird. Columns: repositories, deletions, combined
    log-hazard slope per commit (95\% CI), its $\tau^2$, median frailty variance $\theta$, and the
    share of slope the frailty absorbs. Deaths are censored at 50 commits.
  }
  \centering\small
  \begin{adjustbox}{max width=\linewidth}
  \begin{tabular}{@{}lrrrrrrr@{}}
    \toprule
    Frailty & Repos. & Deaths & Slope & 95\% CI & $\tau^2$ & $\theta$ & Absorbed \\
    \midrule
    no frailty & 118 & 10,226 & $-0.0836$ & $[-0.1054, -0.0619]$ & 0.0098 & -- & -- \\
    shared within file & 118 & 10,226 & $-0.0796$ & $[-0.1013, -0.0580]$ & 0.0096 & 0.001 & 4.8\% \\
    \bottomrule
  \end{tabular}
  \end{adjustbox}
  \end{captionedtable}
\end{table}

%% file: assets/table_thresholds.tex
\begin{table}[H]
  \begin{captionedtable}{tab:thresholds}{
    The deletion-hazard decline survives every choice varied here. Each block moves one choice off
    the published configuration (rewrite floor 5, fraction 0.75, minimum file history
    3 commits, minimum age 1), whose row is \textbf{bold}. \emph{Rewrite-censored}: share
    of deaths the rule removes; slope: $d \log h / da$ per commit with repository-bootstrap 95\% CI;
    \emph{Hazard form}: top two of 5 shapes by $R^2_w$ under the two-stage estimator of
    \autoref{tab:hazardforms}. Rows share one death set and bootstrap seed.
  }
  \centering\small
  \begin{adjustbox}{max width=\linewidth}
  \begin{tabular}{@{}lrrllrlr@{}}
    \toprule
    & & & & \multicolumn{4}{c}{Hazard form (two-stage)} \\
    \cmidrule(l){5-8}
    & & Rewrite- & Log-hazard slope & \multicolumn{2}{c}{Best} & \multicolumn{2}{c}{Runner-up} \\
    \cmidrule(lr){5-6}\cmidrule(l){7-8}
    Setting & Deletions & censored & [95\% CI] & form & $R^2_w$ & form & $R^2_w$ \\
    \midrule
    \multicolumn{8}{@{}l}{\emph{Rewrite fraction of live instructions (floor 5)}} \\
    0.25 & 16,383 & 85.5\% & $-0.050$ $[-0.062, -0.041]$ & \texttt{geometric} & 0.913 & \texttt{weibull} & 0.633 \\
    0.50 & 26,643 & 76.3\% & $-0.034$ $[-0.044, -0.022]$ & \texttt{geometric} & 0.916 & \texttt{weibull} & 0.562 \\
    {\boldmath\textbf{0.75}} & {\boldmath\textbf{38,138}} & {\boldmath\textbf{66.1\%}} & {\boldmath\textbf{$-0.043$ $[-0.054, -0.032]$}} & {\boldmath\textbf{\texttt{geometric}}} & {\boldmath\textbf{0.815}} & {\boldmath\textbf{\texttt{weibull}}} & {\boldmath\textbf{0.291}} \\
    \midrule
    \multicolumn{8}{@{}l}{\emph{Minimum file history (commits)}} \\
    {\boldmath\textbf{3}} & {\boldmath\textbf{38,138}} & {\boldmath\textbf{66.1\%}} & {\boldmath\textbf{$-0.043$ $[-0.054, -0.032]$}} & {\boldmath\textbf{\texttt{geometric}}} & {\boldmath\textbf{0.815}} & {\boldmath\textbf{\texttt{weibull}}} & {\boldmath\textbf{0.291}} \\
    4 & 36,712 & 65.5\% & $-0.038$ $[-0.055, -0.023]$ & \texttt{geometric} & 0.789 & \texttt{weibull} & 0.468 \\
    5 & 35,208 & 65.0\% & $-0.039$ $[-0.056, -0.025]$ & \texttt{geometric} & 0.756 & \texttt{weibull} & 0.322 \\
    \bottomrule
  \end{tabular}
  \end{adjustbox}
  \end{captionedtable}
\end{table}

%% file: sections/10b-appendix-inverse-ifeval.tex
\section{The Inverse-IFEval Testbed}
\label{sec:appendix:testbed}

\subsection{Worlds, roles, and the censored feedback channel}
\label{sec:appendix:worlds}
\label{sec:appendix:loop}
\label{sec:appendix:regime}

\input{assets/table_verifiers.tex}
A world is an IFEval item with at least two verifiers, its objective $o_j$ written once by a stronger model. Each
round the executor answers \AccessL{} sampled task, \AccessK{} sampled hidden constraints score the response, and a
fresh maintainer adds or deletes instructions, never rewording one, so every instruction's age is unambiguous
(\autoref{sec:measurement}). A failure returns its complaint (\autoref{tab:verifiers}), a pass a bare verdict; the
maintainer never learns how many constraints exist, which were checked, or which instruction passed one. Both roles
are \texttt{claude-haiku-4-5}.

\subsection{The comment protocol}
\label{sec:appendix:channels}
Every arm's maintainer receives one system prompt (\autoref{fig:prompts:maintainer}); arms differ only in
what the maintainer writes after
\texttt{\#} and later reads there. The maintainer may add an instruction (one imperative sentence about the
response, never mentioning scores, failures, rounds or history) or delete one by id. Every operation names
the requirement it targets.

In arms that write comments, every operation also carries a one-line comment of at most \LabCommentCap{}
characters in a fixed form. An addition's comment records its
round, one checkable hypothesis about the hidden requirement with its evidence, the failure that prompted it (quoted,
with the round first seen), the observation that would refute it, the hypotheses already ruled out, and the text of
any instruction it replaces; a deletion's records the hypothesis dropped and why.

The harness strips every comment before the executor reads the prompt (\autoref{fig:prompts:executor}). The
\textsc{ShuffledNoise} maintainer reads, in
place of its own comments, those \textsc{PromptComments} wrote in other worlds on constraints of a different type.

\input{assets/fig_prompts.tex}

\FloatBarrier

\subsection{Verifying the reference cover}
\label{sec:appendix:cover}

We verify the reference cover $D_\star$ against the executor itself, with no maintainer in the loop. Dropping one of
its \CoverDirectives{} instructions costs that instruction's own constraint \CoverLooCost{} on average
($z = \CoverLooZ$). The other \CoverExcessShare{} cost nothing: the executor satisfies them unprompted. Removing
them gives the \emph{verified cover} $D_V$, so excess size against $D_\star$ understates excess against the true
minimum.

\subsection{The long-horizon run}
\label{sec:appendix:lab}
\label{sec:appendix:horizon}
\label{sec:appendix:capacity}

\input{assets/table_arms.tex}

\subsection{Inline exposure}
\label{sec:appendix:failures}

The harness strips comments before the executor reads the prompt (\autoref{sec:appendix:channels}). Showing them to
the executor as well costs size and buys no constraint satisfaction: with \InlineExposure{} of executor calls exposed, prompts end
at \InlineExcessPct{} excess against \InlineRefExcessPct{} with comments stripped (\InlineDeltaExcessCI{}, one seed,
\InlineWorlds{} worlds), at a constraint-satisfaction difference of \InlineDeltaSatCI{}.

\subsection{Comment ablations}
\label{sec:appendix:ablations}

\input{assets/table_ablations.tex}

\subsection{Where the comment-enabled arm ends}
\label{sec:appendix:outcomes}

\input{assets/table_outcomes.tex}

\subsection{Earlier protocols}
\label{sec:appendix:pending}
\label{sec:appendix:ladder}

Read together, the failures in \autoref{tab:ladder} motivate the canonical protocol: each comment records both the
maintainer's intention (its hypothesis) and the outcomes that test it (the failure that prompted it, what would refute
it, and what has been ruled out; \autoref{sec:appendix:channels}).

\input{assets/table_ladder.tex}

\subsection{Unprompted rewrites and richer feedback}
\label{sec:appendix:worldrewrites}

The worlds never prompt a rewrite, yet maintainers rewrite wholesale on their own. Under the corpus rule (a round
killing at least \RewriteFracThreshold{} of at least \RewriteMinDirectives{} live instructions), the uncommented
arm of \autoref{tab:arms} rewrites \WorldRewritesControl{} times per world over \HorizonT{} rounds against
\WorldRewritesCommit{} for the comment-enabled arm (paired difference $\WorldRewritesDelta$, 95\pct{} bootstrap CI
$\WorldRewritesDeltaCI$, \HorizonWorlds{} worlds). As corpus rewrites follow growth
(\autoref{sec:rewrite}), \WorldRewritesGrewControl{} of the uncommented rewrites follow three rounds of growth,
\WorldRewritesGrewDelta{} more often than other rounds large enough to be rewritten (95\pct{} CI
$\WorldRewritesGrewDeltaCI$\pp{}, resampling worlds).

Feedback that names the rule leaves the effect in place. When CodeIF's judge names the failing convention rather
than only describing the failure (\autoref{sec:wild}), comments still shrink the prompt by \CodeifNamedSizePp{} of
$|D_\star|$ against the uncommented arm, at a constraint-satisfaction difference of \CodeifNamedSatPp{}
(\CodeifNamedWorlds{} fresh worlds, one seed, \BeyondT{} rounds, the maintainer of \autoref{tab:beyond}).

%% file: assets/table_verifiers.tex
\begin{table}[H]
  \begin{captionedtable}{tab:verifiers}{
    The hidden constraints and the only feedback the maintainer receives, by IFEval family. $n$: (seed, world)
    pairs carrying the type; each world carries two or three. A complaint names a dimension, never a parameter: the
    censoring \S\ref{sec:theory} assumes.
  }
  \centering\small
  \setlength{\tabcolsep}{3pt}
  \begin{tabularx}{\linewidth}{@{}llr>{\raggedright\arraybackslash}X@{}}
    \toprule
    Family & Verifier type & $n$ & Complaint returned on failure \\
    \midrule
    keywords & \texttt{existence} & 69 & the response is missing some required content \\
     & \texttt{forbidden\_words} & 81 & the response contains wording it must not use \\
     & \texttt{frequency} & 75 & certain wording is not used the right number of times \\
     & \texttt{letter\_frequency} & 57 & a certain letter does not appear the right number of times \\
    \addlinespace
    length & \texttt{number\_paragraphs} & 48 & the response is not split into the right number of paragraphs \\
     & \texttt{number\_sentences} & 99 & the response has the wrong number of sentences \\
     & \texttt{number\_words} & 108 & the response's word count is off \\
    \addlinespace
    content & \texttt{number\_placeholders} & 42 & the response needs a different number of bracketed placeholders \\
     & \texttt{postscript} & 36 & the response is missing a postscript \\
    \addlinespace
    format & \texttt{json\_format} & 27 & the response is not valid JSON \\
     & \texttt{multiple\_sections} & 24 & the response is not organized into the expected sections \\
     & \texttt{number\_bullet\_lists} & 54 & the response does not have the right number of bullet lists \\
     & \texttt{number\_highlighted\_sections} & 87 & the response needs a different number of highlighted sections \\
     & \texttt{title} & 36 & the response is missing a properly formatted title \\
    \addlinespace
    case & \texttt{capital\_word\_frequency} & 39 & the number of fully-capitalized words is off \\
     & \texttt{english\_capital} & 42 & the response's letter casing is wrong \\
     & \texttt{english\_lowercase} & 75 & the response's letter casing is wrong \\
    \addlinespace
    punct. & \texttt{no\_comma} & 87 & the response's punctuation breaks a rule \\
    \addlinespace
    start/end & \texttt{end\_checker} & 27 & the response does not end the required way \\
     & \texttt{quotation} & 69 & the response does not begin and end the required way \\
    \bottomrule
  \end{tabularx}
  \end{captionedtable}
\end{table}

%% file: assets/fig_prompts.tex
\begin{figure*}[htbp]
  \scriptsize
  \begin{subfigure}{\linewidth}
  \caption{The maintainer's system prompt, shared by every arm}\label{fig:prompts:maintainer:shared}
  \scriptsize
  {\color{black!70}
  \begin{verbatim}
You maintain a reusable prompt (a numbered list of directives) for an assistant that repeatedly performs
variations of one task whose requirements are hidden. Your objective is to learn the hidden requirements to
optimize the prompt and maximize the LONG-RUN average score.

After each round, the assistant answers using your prompt, and its response is checked against a random sample of
the hidden requirements; the scores (and possibly a description of any failures) are reported to you. You are NOT
told how many requirements exist, which were sampled, or which directive (if any) satisfied a passing check.
Because each round checks only some requirements, a round without a failure is not necessarily evidence that a
directive is no longer needed.

When comments are shown (after `#`), they are earlier maintainers' notes on why each directive was added, and
your only record of earlier rounds. Before adding or deleting anything, read them all and incorporate their
reasoning into yours.

You may change the prompt with these operations:
- add: append a directive, one imperative sentence addressed to the assistant about what the RESPONSE must do. It
must NEVER mention scores, failures, rounds, or history. Adding usually makes sense when the evidence so far
shows a failure that no live directive seems to address.
- delete: remove a directive by its id. Deleting usually makes sense when the evidence so far suggests a
directive isn't doing its job or is causing a failure. You cannot reword a directive; to change one, delete it
and add a new one.
- {arm_bullet}

Return only the operations. An empty ops list keeps the prompt unchanged.
\end{verbatim}
  }
  \end{subfigure}

  \begin{subfigure}{\linewidth}
  \caption{\texttt{\{arm\_bullet\}} for every arm but \textsc{None}}\label{fig:prompts:maintainer:commenting}
  \scriptsize
  {\color{black!70}
  \begin{verbatim}
With each add or delete, supply `cause` (one short phrase naming the requirement the directive targets, e.g. "no
commas allowed") and `comment` (one line of at most 1,024 characters), in this form:

    add:    r<N> | hypothesis: <claim> because <evidence> | for: "<failure>" since r<M> | refuted if:
    <observation> | ruled out: <claims> | restores: "<directive>"
    delete: hypothesis: <claim> | dropped because: <evidence>

  where
    N             = one more than the highest r<N> shown (r1 if none)
    <claim>       = one concrete, checkable belief about the hidden requirement (e.g. "the requirement is exactly
    4 sections"); on a delete, the belief the deleted directive encoded
    <evidence>    = what in the feedback, the response, or the ruled-out claims supports this
    <failure>     = the feedback that prompted it, quoted verbatim
    M             = the round that failure was first seen: copy it from the comments of earlier directives on the
    same failure, else N
    <observation> = what would show <claim> wrong, usually: "<failure>" recurs while this is live
    <claims>      = claims already refuted for this failure, copied forward from the comments of directives you
    delete
    <directive>   = the exact text of a deleted directive this one replaces, so it can be restored
  Omit any field that would be empty; never write "none".

  Example: r6 | hypothesis: the requirement is exactly 4 sections because 3 and 5 both failed and the response's
  4-part draft drew no section complaint | for: "the response needs a different number of highlighted sections"
  since r2 | refuted if: that failure recurs while this is live | ruled out: 3 sections, 5 sections | restores:
  "Use exactly 5 highlighted sections."
\end{verbatim}
  }
  \end{subfigure}

  \begin{subfigure}{\linewidth}
  \caption{\texttt{\{arm\_bullet\}} for \textsc{None}}\label{fig:prompts:maintainer:none}
  \scriptsize
  {\color{black!70}
  \begin{verbatim}
With each add or delete, supply `cause` (one short phrase naming the requirement the directive targets, e.g. "no
commas allowed").
\end{verbatim}
  }
  \end{subfigure}

  \caption{
    Every prompt the loop issues, verbatim, with line breaks set for this page. \textbf{(a)} The maintainer's system
    prompt (\autoref{tab:access}), with \textbf{(b)} or \textbf{(c)} in its \texttt{\{arm\_bullet\}}; the commenting
    arms are then shown nothing, another world's comments, or their own. \textbf{(d)} The executor's
    comment-stripped and \textbf{(e)} the maintainer's input in \textsc{PromptComments}' last round on world 20 (3
    instructions; arm median 2).
  }
  \label{fig:prompts:maintainer}
\end{figure*}

\begin{figure*}[htbp]
  \ContinuedFloat
  \scriptsize
  \begin{subfigure}{\linewidth}
  \caption{The executor's system prompt and user message in one round}\label{fig:prompts:executor}
  \scriptsize
  {\color{black!70}
  \begin{verbatim}
Follow ALL of these directives in your response:
1. Begin your response with a title on its own line in the format TITLE: [content], then provide the main content
on subsequent lines.
2. Ensure all letters in the main content (excluding the title) are lowercase.
3. Ensure the letter 'e' appears exactly 8 times in your response.

--- user message ---

Write a haiku about Ireland in the style of Basho.
\end{verbatim}
  }
  \end{subfigure}

  \begin{subfigure}{\linewidth}
  \caption{The maintainer's user message that round: its commented instructions and the round's scores}\label{fig:prompts:maintainer:round}
  \scriptsize
  {\color{black!70}
  \begin{verbatim}
## Current prompt directives
[d16] Begin your response with a title on its own line in the format TITLE: [content], then provide the main
content on subsequent lines.  # r12 | hypothesis: the title must be on a separate line from content because the
format-only directive failed; structural separation may be the missing requirement | for: 'the response is
missing a properly formatted title' since r5 | refuted if: title failure recurs while this is live | ruled out:
square brackets are the required format
[d18] Ensure all letters in the main content (excluding the title) are lowercase.  # r13 | hypothesis: the
requirement is that main content must be all lowercase because the failure 'the response's letter casing is
wrong' appeared with mixed case in the response (TITLE: Irish Mist  Green isle emerges...) where Green and Soft
are capitalized in the body | for: 'the response's letter casing is wrong' since r13 | refuted if: this failure
recurs while this is live
[d19] Ensure the letter 'e' appears exactly 8 times in your response.  # r14 | hypothesis: the requirement
targets 'e' at exactly 8 occurrences because the current response has 8 e's and still fails, suggesting either
the count or the scope is wrong; if count is 8, the failure must be due to scope (perhaps excluding title, or
including/excluding whitespace differently) | for: 'a certain letter does not appear the right number of times'
since r2 | refuted if: this failure recurs while this is live | ruled out: 6 for 'e', 7 for 'e', 10 for 'e', 15
for 'e' | restores: 'Ensure the letter \'e\' appears exactly 7 times in your response.'

## This round's scores
task 0: a requirement (the response's letter casing is wrong) FAILED
task 0: a requirement (the response is missing a properly formatted title) FAILED
task 0 response (truncated): TITLE: irish mist  green hills bow and bend, emerald dreams drift through mist—
silence holds the stone.
\end{verbatim}
  }
  \end{subfigure}

  \caption[]{
Continued.
  }
\end{figure*}

%% file: assets/table_arms.tex
\begin{table}[H]
  \begin{captionedtable}{tab:arms}{
    \textsc{PromptComments} cuts excess size at parity satisfaction (\HorizonT{} rounds, 184 worlds, one seed).
    \textsc{FixedNoise} reads generic comments drawn from a fixed pool. Excess is $|D_T|/|D_\star|{-}1$ at $|D_\star|{=}2.2$; satisfaction is the share of scored constraints passed,
    at $t{=}0$ and over the last 3 rounds ($t{=}T$). \textbf{Bold}: best arm, CIs disjoint; $\pm$: half a 95\pct{}
    bootstrap CI (asymmetric for levels).
  }
  \centering\small
  \setlength{\tabcolsep}{4pt}
  \begin{tabular*}{\textwidth}{@{\extracolsep{\fill}}lrrrr@{}}
    \toprule
    & \multicolumn{2}{c}{Instruction count} & \multicolumn{2}{c}{Constraint satisfaction} \\
    \cmidrule(lr){2-3}\cmidrule(lr){4-5}
    Arm & $|D_T|$ & Excess (\pct) & $t{=}T$ (\pct) & $t{=}0$ (\pct) \\
    \midrule
    \textsc{None} & 6.6 & $+211.3 \pm 105.3$ & $44.0 \pm 4.9$ & $23.9 \pm 3.9$ \\
    \textsc{FixedNoise} & 5.3 & $+147.9 \pm 79.9$ & $42.6 \pm 4.9$ & $22.3 \pm 4.2$ \\
    \textsc{PromptComments} & 2.2 & {\boldmath$+1.4 \pm 22.1$} & $44.0 \pm 4.7$ & $21.7 \pm 3.9$ \\
    \bottomrule
  \end{tabular*}
  \end{captionedtable}
\end{table}

%% file: assets/table_ablations.tex
\begin{table}[H]
  \begin{captionedtable}{tab:ablations}{
    The schema carries \textsc{PromptComments}' size reduction over \textsc{None}: a free-form narrative in its place
    (\textsc{Freeform}) gives back \AblationFreeformGivebackPct{} of it; removing a single field raises excess, though
    no single removal resolves from zero at one seed.
    Ablations share 64 worlds (\AccessT{} rounds, one seed).
    Excess $|D_T|/|D_\star|{-}1$ and satisfaction over the last 3 rounds ($t{=}T$); $\Delta$: row $-$
    \textsc{PromptComments} (pp); $\pm$: 95\pct{} bootstrap half-width over worlds.
  }
  \centering\small
  \setlength{\tabcolsep}{4pt}
  \begin{tabularx}{\linewidth}{@{}>{\raggedright\arraybackslash}Xrrrr@{}}
    \toprule
    & \multicolumn{2}{c}{Excess vs.\ $D_\star$} & \multicolumn{2}{c}{Satisfaction, $t{=}T$} \\
    \cmidrule(lr){2-3}\cmidrule(lr){4-5}
    Arm & (\pct) & $\Delta$ (pp) & (\pct) & $\Delta$ (pp) \\
    \midrule
    \textsc{None} & $+84.1 \pm 63.3$ & $+90.1 \pm 60.8$ & $38.3 \pm 7.8$ & $-1.6 \pm 6.1$ \\
    \textsc{Freeform} & $+83.6 \pm 51.6$ & $+89.6 \pm 47.8$ & $38.5 \pm 7.6$ & $-1.3 \pm 5.9$ \\
    \textsc{PromptComments} & $-6.0 \pm 15.2$ & -- & $39.8 \pm 7.4$ & -- \\
    \quad w/o subsequent performance & $+15.9 \pm 22.4$ & $+21.9 \pm 25.9$ & $39.6 \pm 7.9$ & $-0.3 \pm 6.9$ \\
    \quad w/o ruled-out hypotheses & $+10.2 \pm 26.6$ & $+16.1 \pm 18.4$ & $37.0 \pm 8.2$ & $-2.9 \pm 4.9$ \\
    \quad w/o restore quote & $-5.2 \pm 17.4$ & $+0.8 \pm 21.4$ & $37.2 \pm 7.8$ & $-2.6 \pm 6.3$ \\
    \bottomrule
  \end{tabularx}
  \end{captionedtable}
\end{table}

%% file: assets/table_outcomes.tex
\begin{table}[H]
  \begin{captionedtable}{tab:outcomes}{
    At every final size, the commented arm's satisfaction matches or exceeds the uncommented arm's.
    \autoref{tab:access}'s worlds (\AccessWorlds{}, \AccessSeeds{} seeds, \AccessT{} rounds) grouped by
    \textsc{PromptComments}' final size against $|D_\star|$, emptied ones by cause. Share uses each arm's own final
    size; excess is $|D_T|/|D_\star|{-}1$; satisfaction is over the last 3 rounds.
  }
  \centering\small
  \setlength{\tabcolsep}{4pt}
  \begin{tabularx}{\linewidth}{@{}>{\raggedright\arraybackslash}Xrrrrrr@{}}
    \toprule
    & \multicolumn{2}{c}{Share (\pct)} & \multicolumn{2}{c}{Excess vs.\ $D_\star$ (\pct)} & \multicolumn{2}{c}{Sat.\ (\pct)} \\
    \cmidrule(lr){2-3}\cmidrule(lr){4-5}\cmidrule(lr){6-7}
    Final state & \shortstack[r]{\textsc{Prompt}\\\textsc{Comments}} & \textsc{None} & \shortstack[r]{\textsc{Prompt}\\\textsc{Comments}} & \textsc{None} & \shortstack[r]{\textsc{Prompt}\\\textsc{Comments}} & \textsc{None} \\
    \midrule
    At $|D_\star|$ & $38.5$ & $29.2$ & $+0.0$ & $+28.2$ & $33.8$ & $33.3$ \\
    Below $|D_\star|$ & $35.9$ & $20.3$ & $-48.8$ & $-10.6$ & $51.4$ & $49.8$ \\
    Above $|D_\star|$ & $15.1$ & $35.4$ & $+151.7$ & $+240.2$ & $50.6$ & $37.9$ \\
    Emptied & $10.4$ & $15.1$ & $-100.0$ & $-42.5$ & $36.7$ & $35.8$ \\
    \quad unlearnable by any arm & $4.7$ & -- & $-100.0$ & $-66.7$ & $42.6$ & $38.9$ \\
    \quad only learned by others & $3.1$ & -- & $-100.0$ & $-8.3$ & $36.1$ & $38.9$ \\
    \quad deleted after previously learned & $2.6$ & -- & $-100.0$ & $-40.0$ & $26.7$ & $26.7$ \\
    \bottomrule
  \end{tabularx}
  \end{captionedtable}
\end{table}

%% file: assets/table_ladder.tex
\begin{table}[H]
  \begin{captionedtable}{tab:ladder}{
    Earlier protocols, in the order tried, each from its own run. Excess is $|D_T|/|D_\star|{-}1$, Sat.\ is
    satisfaction over the last 3 rounds, Empty the share of prompts ending empty. Uncommented controls' excess varies
    with the worlds drawn ($N$), $+56.2\%$ to $+84.1\%$. \textsc{Freeform} exceeds its own control ($+70.0\%$ against
    $+56.2\%$): persisting actions without outcomes is worse than persisting nothing.
  }
  \centering\small
  \begin{adjustbox}{max width=\linewidth}
  \begin{tabular}{@{}l>{\raggedright\arraybackslash}p{0.19\textwidth}>{\raggedright\arraybackslash}p{0.36\textwidth}rrrr@{}}
    \toprule
    & Persists & Failure mode & $N$ & Excess & Sat. & Empty \\
    \midrule
    \textsc{Freeform} & free-text strategy & actions without outcomes: successors extend failing hypotheses & 184 & $+70.0\%$ & 0.377 & 2.7\% \\
    \textsc{Ledger} & confirmed / falsified / open note & suppresses additions, never prunes: no per-instruction record & 184 & $+42.8\%$ & 0.407 & 4.9\% \\
    \textsc{Tally} & note + evidence per instruction & nothing over the note: a bounded note saturates & 64 & $+57.8\%$ & 0.411 & 3.1\% \\
    \textsc{Window} & note + 5-round digest & reaches cover by emptying prompts; the maintainer does not write the digest & 184 & $-1.9\%$ & 0.361 & 17.9\% \\
    \textsc{Calibrated} & digest + observed / guess / evidence & hedging suppresses additions, costs satisfaction & 64 & $-4.7\%$ & 0.333 & 14.1\% \\
    \bottomrule
  \end{tabular}
  \end{adjustbox}
  \end{captionedtable}
\end{table}

%% file: sections/10d-appendix-live.tex
\section{The Real-Repository Pruning Experiment}
\label{sec:appendix:live}

This appendix details \autoref{sec:live}'s selection, reconstruction, pruning and scoring.

\paragraph{Selection.} We score \LiveJointPairs{} task--instruction pairs from \LiveTasks{} tasks on \LiveFiles{}
real agentic prompts. Of \LiveEvents{} wholesale rewrites (\autoref{sec:measurement}) in \LiveEventRepos{}
repositories, we keep each repository's earliest with at least five file versions before it and five task commits
after, a post-rewrite file $D_{\text{post}}$ of at least five instructions, and under a quarter of $D_{\text{pre}}$
reworded rather than dropped, so the rewrite removes instructions a prune could also remove. We score files whose
$D_{\text{pre}}$ holds under \LiveMaxDirectives{} instructions, so each file's history fits one comment-writing
pass, and tasks on which $D_{\text{pre}}$'s own response passes at least one instruction.
\LiveUnsignedWorlds{} of the \LiveFiles{} rewrites carry no coding agent's signature (a \texttt{Co-Authored-By}
trailer or bot author from Claude Code, Codex, Cursor, Copilot and others).

\paragraph{Task reconstruction.} Each task is a commit co-authored by Claude Code after the rewrite, while
$D_{\text{post}}$ was in force.
We keep commits with an informative message that leave the agentic prompt alone and touch at least one file other
than lockfiles, binaries and agent configuration, one per normalized subject, and score each file on its oldest
such commits.
The executor (\texttt{claude-haiku-4-5}) receives the commit message and the files it touched at its parent (at
most four), and returns a commit message, commands and a diff.

\paragraph{Comment reconstruction.} A reasoning model reads the repository's history up to the commit before the
rewrite (its last 256 commits, selected diffs, a directory tree and a documentation index) and writes each
$D_{\text{pre}}$ instruction a comment: why it exists, what the history shows of it since, and what a maintainer
should do next. Nothing at or after the rewrite is visible.

\paragraph{Pruning.} One maintainer call per arm (\texttt{claude-sonnet-5}, one seed) prunes $D_{\text{pre}}$ with those comments
(\textsc{PromptComments}) or without (\textsc{None}), under \autoref{sec:wild}'s CodeIF update rule, system
prompt and edit operations; the executor answers each task under both prunes, $D_{\text{pre}}$ and
$D_{\text{post}}$.

\paragraph{Two-judge scoring.} Two reasoning judges (\texttt{gpt-5.6-luna} and \texttt{gemini-3.8-flash}) mark
every $D_{\text{post}}$ instruction applicable and passed or failed.
Each judge scores every arm on the pairs it deems applicable to any arm's response, a skipped
instruction counting as unmet; a file's constraint satisfaction is its mean per-task passed share, weighted by pairs
across files and judges (\autoref{tab:live}).

\paragraph{Judge validity.} The judges never see the arm: each call shows the task, the $D_{\text{post}}$
instructions and one response, never the pruned file, its comments or the arm's name.
We hand-labelled task--instruction pairs from \textsc{PromptComments} and \textsc{None} responses, blind to arm, and
compared the labels with both judges on the \LiveJointAuditItems{} labelled pairs where the judges disagree.
Those disagreements turn mostly on whether an instruction applies, not whether it was followed, and the labels side with \texttt{gpt-5.6-luna} on \LiveJointAuditLuna{} and with
\texttt{gemini-3.8-flash} on \LiveJointAuditSecond{}.